\documentclass[letterpaper]{article} 
\usepackage{aaai25}  
\usepackage{times}  
\usepackage{helvet}  
\usepackage{courier}  
\usepackage[hyphens]{url}  
\usepackage{graphicx} 
\usepackage{natbib}  
\usepackage{caption} 
\usepackage{algorithm}
\usepackage{algorithmic}
\usepackage{amsmath,amssymb,amsfonts,amsthm,romannum}
\theoremstyle{definition}
\newtheorem{theorem}{Theorem}
\newtheorem*{remark}{Remark}
\newtheorem{lemma}{Lemma}

\newtheorem{corollary}{Corollary}
\newtheorem{assumption}{Assumption}
\def\ie{{\it i.e.,\ \/}}

\title{Doubly-Bounded Queue for Constrained Online Learning:
	\\Keeping Pace with Dynamics of Both Loss and Constraint}
\author{
    Juncheng Wang\thanks{Corresponding author, jcwang@comp.hkbu.edu.hk},
    Bingjie Yan,
    Yituo Liu
}
\affiliations{
    Department of Computer Science, Hong Kong Baptist University
}

\begin{document}

\maketitle

\begin{abstract}
	We consider online convex optimization with time-varying constraints and conduct performance analysis using two stringent metrics: dynamic regret with respect to the online solution benchmark, and hard constraint violation that does not allow any compensated violation over time. We propose an efficient algorithm called Constrained Online Learning with Doubly-bounded Queue (COLDQ), which introduces a novel virtual queue that is both lower and upper bounded, allowing tight control of the constraint violation without the need for the Slater condition. We prove via a new Lyapunov drift analysis that COLDQ achieves $\mathcal{O}(T^\frac{1+V_x}{2})$ dynamic regret and $\mathcal{O}(T^{V_g})$ hard constraint violation, where $V_x$ and $V_g$ capture the dynamics of the loss and constraint functions. For the first time, the two bounds smoothly approach to the best-known $\mathcal{O}(T^\frac{1}{2})$ regret and $\mathcal{O}(1)$ violation, as the dynamics of the losses and constraints diminish. For strongly convex loss functions, COLDQ matches the best-known $\mathcal{O}(\log{T})$ static regret while maintaining the $\mathcal{O}(T^{V_g})$ hard constraint violation. We further introduce an expert-tracking variation of COLDQ, which achieves the same performance bounds without any prior knowledge of the system dynamics. Simulation results demonstrate that COLDQ outperforms the state-of-the-art approaches.
\end{abstract}

\section{Introduction}
\label{sec:1}

In many online learning applications, optimization losses and constraints are dynamic over time. Online Convex Optimization (OCO) \cite{BK-S.Shwartz'12, BK-E.Hazan'16}, as the intersection of learning, optimization, and game, is a vital framework for solving online learning problems under uncertainty. It has broad applications such as advertisement placement \cite{APP:Balseiro-ICML'20}, load balancing \cite{APP:Hsu-OR'21}, network virtualization \cite{APP:M.Shi-TON'21}, and resource allocation \cite{DTCOCO-TON}.

In the standard OCO setting, a learner selects online decisions from a known convex set to minimize a sequence of time-varying convex loss functions. The information of each loss function, however, is only revealed to the learner after the decision has been made. Given this lack of current information, the objective of the learner becomes to minimize the \textit{regret}, which is the accumulated difference between the losses incurred by their online decisions and those of some benchmark solutions. \citet{OCO-STC:M.Zinkevich-ICML'2003} considered both \textit{static} regret to an offline benchmark and \textit{dynamic} regret to an online benchmark. The proposed online projected gradient descent algorithm provided a dynamic regret bound that \textit{smoothly} approaches to $\mathcal{O}(T^\frac{1}{2})$ static regret, as the accumulated variation of the loss functions reduces, \ie the OCO algorithm keeps pace with the dynamics of the losses.

The projection operation to strictly satisfy the constraints at each time can incur heavy computation. Furthermore, in many applications, the online decisions are subject to constraints that are allowed to be violated at certain time slots. \citet{SLTC:M.Mahdavi-JMLR'12} initiated the study on OCO with \textit{soft constraint violation}, which measures the amount of compensated violations over time. In contrast, with a goal to limit the instantaneous violation, \citet{CLTC:J.Yuan-NIPS'18} introduced a stronger notion of \textit{hard} constraint violation that does not allow any compensated violation over time. For \textit{fixed} constraints, the best-known soft and hard constraint violation bounds are both $\mathcal{O}(1)$ \cite{SLTC:H.Yu-JMLR'20,CLTC:H.Guo-NIPS'22}.

Most existing works on OCO with \textit{time-varying} constraints focused on the static regret \cite{SLTC:H.Yu-NIPS'17,SLTC:X.Wei-SIGMETRICS'20,DLTC:X.Cao-TAC'21,CLTC:A.Sinha-NIPS'24}. Dynamic regret for time-varying constrained OCO was more recently studied  \cite{DLTC:T.Chen-TSP'17,DLTC:X.Cao-TAC'19,DLTC:Q.Liu-SIGMETRICS'22,CLTC:H.Guo-NIPS'22,DCLTC:X.Yi-TAC'23,DTCOCO-TON}. As the accumulated variation of the constraint functions reduces, the best-known soft and hard constraint violation bounds for time-varying constraints approach to $\mathcal{O}(T^\frac{1}{2})$ and $\mathcal{O}(T^\frac{1}{2}\log T)$, respectively \cite{DTCOCO-TON, CLTC:A.Sinha-NIPS'24}. However, none of the constraint violation bound recovers the best-known $\mathcal{O}(1)$ violation for fixed constraints, \ie the constrained OCO algorithms do \textit{not} keep pace with the dynamics of the constraints.

The above discrepancies motivate us to pose the following key question: \textit{Can a constrained OCO algorithm provide a dynamic regret bound and a constraint violation bound that smoothly approach to the best-known $\mathcal{O}(T^\frac{1}{2})$ regret and $\mathcal{O}(1)$ violation, respectively, as the dynamics of the losses and constraints diminish?} Our answer is yes.


\textbf{Contributions.} We summarize our contributions below.

\begin{itemize}
	
	\item We propose an effective algorithm named \underline{C}onstrained \underline{O}nline \underline{L}earning with \underline{D}oubly-bounded \underline{Q}ueue (COLDQ) for tackling OCO problems with time-varying constraints. Existing virtual-queue-based approaches rely on either a lower or an upper bound of the virtual queue to bound the constraint violation. In contrast, we introduce a novel virtual queue that enforces both a lower and an upper bound, without the commonly assumed Slater condition, to strictly control the constraint violation.
	
	\item We analyze the performance of COLDQ via a new Lyapunov drift design that leverages both the lower and upper bounds of the virtual queue. We show that COLDQ provides $\mathcal{O}(T^\frac{1+V_x}{2})$ dynamic regret and $\mathcal{O}(T^{V_g})$ hard constraint violation, where $V_x$ and $V_g$ capture the dynamics of the losses and constraints (see definitions in (\ref{eq:Vx}) and (\ref{eq:Vg})). For the first time, the two bounds smoothly approach to the best-known $\mathcal{O}(T^\frac{1}{2})$ regret and $\mathcal{O}(1)$ violation as $V_x\to0$ and $V_g\to0$.
	
	\item When the loss functions are strongly convex, we show that COLDQ matches the best-known $\mathcal{O}(\log{T})$ static regret, while maintaining the $\mathcal{O}(T^{V_g})$ hard constraint violation. We further propose a variation of COLDQ with expert tracking that can achieve the same $\mathcal{O}(T^\frac{1+V_x}{2})$ dynamic regret and $\mathcal{O}(T^{V_g})$ hard constraint violation, without any prior knowledge about the system dynamics.
	
	\item We conduct experiments to evaluate the practical performance of COLDQ on various applications involving both time-varying and fixed constraints. Numerical results confirm the effectiveness of COLDQ over the state-of-the-art approaches.
	
\end{itemize}


\begin{table*}[t]
	\centering
	\small{
			\begin{tabular}{c|c|c|c}\hline
				Reference & Loss Function & Static Regret, Hard Constraint Violation
				& Dynamic Regret, Hard Constraint Violation\\\hline\hline
				\citet{CLTC:H.Guo-NIPS'22}&  Convex & $\mathcal{O}(T^\frac{1}{2})$, \quad$\mathcal{O}(T^\frac{3}{4})$
				& $\mathcal{O}(T^{\frac{1}{2}+V_x})$, \quad$\mathcal{O}(T^\frac{3}{4})$\\\hline
				\citet{DCLTC:X.Yi-TAC'23} & Convex & $\mathcal{O}(T^{\frac{1}{2}})$, \quad$\mathcal{O}(T^\frac{3}{4})$
				& N/A\\\hline
				\citet{CLTC:A.Sinha-NIPS'24}& Convex & $\mathcal{O}(T^\frac{1}{2})$, \quad $\mathcal{O}(T^\frac{1}{2}\log T)$ & N/A \\\hline
				COLDQ (this work) & Convex & $\mathcal{O}(T^\frac{1}{2})$, \quad$\mathcal{O}(T^{V_g})$
				& $\mathcal{O}(T^\frac{1+V_x}{2})$,\quad $\mathcal{O}(T^{V_g})$\\\hline
				\citet{CLTC:H.Guo-NIPS'22}&  Strongly convex & $\mathcal{O}(\log T)$, \quad$\mathcal{O}(T^\frac{1}{2}(\log T)^\frac{1}{2})$
				& $\mathcal{O}(T^{\frac{1}{2}+V_x})$, \quad$\mathcal{O}(T^\frac{1}{2}(\log T)^\frac{1}{2})$\\\hline
				\citet{DCLTC:X.Yi-TAC'23} & Strongly convex & $\mathcal{O}(T^c)$, \quad$\mathcal{O}(T^{1-\frac{c}{2}})$ & N/A\\\hline
				\citet{CLTC:A.Sinha-NIPS'24}& Strongly convex & $\mathcal{O}(\log T)$, \quad $\mathcal{O}(T^\frac{1}{2}(\log T)^\frac{1}{2})$ & N/A \\\hline
				COLDQ (this work) & Strongly convex &$\mathcal{O}(\log T)$, \quad$\mathcal{O}(T^{V_g})$
				&$\mathcal{O}(T^\frac{1+V_x}{2})$,\quad $\mathcal{O}(T^{V_g})$\\\hline
			\end{tabular}
	}
	\caption{Performance bounds for time-varying constraints ($V_g>0$).}\label{tab:1}
\end{table*}

\begin{table*}[t]
	\centering
	\small{
			\begin{tabular}{c|c|c|c}\hline 
				Reference & Loss Function & Static Regret, Hard Constraint Violation
				& Dynamic Regret, Hard Constraint Violation\\\hline\hline
				\citet{CLTC:X.Yi-ICML'21}  & Convex & $\mathcal{O}(T^\frac{1}{2})$,\quad $\mathcal{O}(T^\frac{1}{4})$
				& $\mathcal{O}(T^\frac{1+V_x}{2})$, \quad $\mathcal{O}(T^\frac{1}{2})$\\\hline
				\citet{CLTC:H.Guo-NIPS'22} &  Convex & $\mathcal{O}(T^\frac{1}{2})$, \quad$\mathcal{O}(1)$
				& $\mathcal{O}(T^\frac{1+V_x}{2})$, \quad$\mathcal{O}(\log T)$\\\hline
				\citet{CLTC:A.Sinha-NIPS'24}& Convex & $\mathcal{O}(T^\frac{1}{2})$, \quad $\mathcal{O}(T^\frac{1}{2}\log T)$ & N/A \\\hline
				COLDQ (this work) & Convex & $\mathcal{O}(T^\frac{1}{2})$, \quad$\mathcal{O}(1)$ &
				$\mathcal{O}(T^\frac{1+V_x}{2})$,\quad $\mathcal{O}(1)$\\\hline
				\citet{CLTC:X.Yi-ICML'21}  & Strongly convex & $\mathcal{O}(\log T)$, \quad$\mathcal{O}(\log T)$
				& $\mathcal{O}(T^\frac{1+V_x}{2})$, \quad$\mathcal{O}(T^\frac{1}{2})$\\\hline
				\citet{CLTC:H.Guo-NIPS'22}&  Strongly convex & $\mathcal{O}(\log T)$, \quad$\mathcal{O}(1)$
				& $\mathcal{O}(T^\frac{1+V_x}{2})$, \quad$\mathcal{O}(\log T)$\\\hline
				\citet{CLTC:A.Sinha-NIPS'24}& Strongly convex & $\mathcal{O}(\log T)$, \quad $\mathcal{O}(T^\frac{1}{2}(\log T)^\frac{1}{2})$ & N/A \\\hline				
				COLDQ (this work) & Strongly convex & $\mathcal{O}(\log T)$, \quad$\mathcal{O}(1)$
				&$\mathcal{O}(T^\frac{1+V_x}{2})$,\quad $\mathcal{O}(1)$\\\hline
			\end{tabular}
	}
			\caption{Performance bounds for fixed constraints ($V_g=0$).}\label{tab:2}
\end{table*}

\section{Related Work}
\label{sec:2}

\subsection{OCO with Fixed Constraints}

The seminal OCO work \cite{OCO-STC:M.Zinkevich-ICML'2003} achieved $\mathcal{O}(T^\frac{1}{2})$ static regret and a more meaningful $\mathcal{O}(T^\frac{1+V_x}{2})$ dynamic regret. For strongly convex loss functions, \citet{OCO-STC:E.Hazan-ML'2007} further improved the static regret bound to $\mathcal{O}(\log{T})$. Dynamic regret has gained increased attention in subsequent OCO works \cite{DReg:E.C.Hall-JSTSP'15, DReg:Jadbabaie-AISTATS'15, DReg:L.Zhang-NeurIPS'18, DReg:Nima-ICML'20}. Some further improvements in the dynamic regret have been achieved by exploiting the strong convexity and smoothness properties \cite{DReg:A.Mokhatari-CDC'16, DReg:L.Zhang-NeurIPS'17, DReg:L.Zhang-L4DC'21}. These works all used projection operations to \textit{strictly} satisfy the constraints at each time.

To reduce the computational complexity incurred by the projection operation, \citet{SLTC:M.Mahdavi-JMLR'12} relaxed the complicated short-term constraints to \textit{long-term} constraints, which need to be satisfied in the time-averaged manner. The proposed saddle-point-type algorithm achieved $\mathcal{O}(T^\frac{1}{2})$ static regret and $\mathcal{O}(T^\frac{3}{4})$ constraint violation. Subsequently, \citet{SLTC:R.Jenatton-ICML'16} provided a trade-off between $\mathcal{O}(T^{\max\{c,1-c\}})$ static regret and $\mathcal{O}(T^{1-\frac{c}{2}})$ constraint violation. For constraints satisfying the Slater condition, which excludes equality constraints, the virtual-queue-based algorithm \cite{SLTC:H.Yu-JMLR'20} reached $\mathcal{O}(T^\frac{1}{2})$ static regret and the best-known $\mathcal{O}(1)$ constraint violation. These works all adopted the \textit{soft} constraint violation that allows \textit{compensated} violations over time.

In contrast,  \citet{CLTC:J.Yuan-NIPS'18} aimed at limiting the \textit{instantaneous} constraint violation and considered a stronger notion of \textit{hard} constraint violation, which does not allow any compensated violation over time. The proposed online algorithm obtained $\mathcal{O}(T^{\max\{c,1-c\}})$ static regret and $\mathcal{O}(T^{1-\frac{c}{2}})$ violation. The online algorithm in \cite{CLTC:X.Yi-ICML'21} provided $\mathcal{O}(T^\frac{1+V_x}{2})$ dynamic regret and $\mathcal{O}(T^\frac{1}{2})$ hard constraint violation.

\subsection{OCO with Time-Varying Constraints}

For OCO problems with stochastic constraints, \citet{SLTC:H.Yu-NIPS'17} proposed a virtual-queue-based algorithm and achieved $\mathcal{O}(T^\frac{1}{2})$ expected static regret and $\mathcal{O}(T^\frac{1}{2})$ expected soft constraint violation, under the Slater condition. Similar $\mathcal{O}(T^\frac{1}{2})$ performance guarantees were obtained under a weaker assumption on the Lagrangian multiplier \cite{SLTC:X.Wei-SIGMETRICS'20}. For time-varying constraints with unknown statistics, \citet{DLTC:X.Cao-TAC'21} reached $\mathcal{O}(T^\frac{1}{2})$ static regret and $\mathcal{O}(T^\frac{3}{4})$ soft constraint violation.

The modified saddle-point-type algorithm \cite{DLTC:T.Chen-TSP'17} attained $\mathcal{O}(T^{\max\{\frac{1+V_x}{2},\frac{1+V_g}{2}\}})$ dynamic regret and $\mathcal{O}(T^{\max\{1-V_x,1-V_g\}})$ soft constraint violation, when the Slater constant is sufficiently large. Another saddle-point-type algorithm \cite{DLTC:X.Cao-TAC'19} achieved $\mathcal{O}(T^\frac{1+V_x}{2})$ dynamic regret and $\mathcal{O}(T^\frac{3+V_x}{4})$ soft constraint violation. \citet{DLTC:Q.Liu-SIGMETRICS'22} proposed a virtual-queue-based algorithm and obtained $\mathcal{O}(T^\frac{1+V_x}{2})$ dynamic regret and $\mathcal{O}(T^{\max\{\frac{3}{4},V_g\}})$ soft constraint violation without the Slater condition. The delay-tolerant algorithm in \cite{DTCOCO-TON} provided $\mathcal{O}(T^{\max\{\frac{1+V_x}{2},V_g\}})$ dynamic regret and $\mathcal{O}(T^{\max\{\frac{1-V_x}{2},V_g\}})$ soft constraint violation under the Slater condition. Unfortunately, as the dynamics of the loss and constraint functions decrease, \ie $V_x\to 0$ and $V_g\to 0$, none of the above \textit{soft} constraint violation bounds approaches to $\mathcal{O}(1)$.

For fixed constraints, \citet{CLTC:H.Guo-NIPS'22} provided the best-known $\mathcal{O}(1)$ hard constraint violation , and was able to keep the $\mathcal{O}(T^\frac{1}{2})$ static regret. For time-varying constraints, \citet{CLTC:H.Guo-NIPS'22} provided $\mathcal{O}(T^\frac{3}{4})$ violation and $\mathcal{O}(T^{\frac{1}{2}+V_x})$ dynamic regret. \citet{DCLTC:X.Yi-TAC'23} achieved $\mathcal{O}(T^\frac{1}{2})$ static regret and $\mathcal{O}(T^\frac{3}{4})$ hard constraint violation under the distributed setting. \citet{CLTC:A.Sinha-NIPS'24} achieved the current best $\mathcal{O}(T^\frac{1}{2}\log T)$ hard constraint violation and $\mathcal{O}(T^\frac{1}{2})$ static regret. Unfortunately still, none of the above \textit{hard} constraint violation bounds smoothly approaches to $\mathcal{O}(1)$ as the system dynamics reduce.

\textbf{Comparisons.} In Tables~\ref{tab:1} and \ref{tab:2}, we compare the performance bounds of COLDQ with the most relevant prior works. The comparison demonstrates that \textit{COLDQ keeps pace with the dynamics of both the losses and constraints}. Below are a few points we would like to highlight.

\begin{itemize}
	
	\item For time-varying constraints and convex loss functions, COLDQ improves upon the current best $\mathcal{O}(T^\frac{1}{2}\log T)$ hard constraint violation bound \cite{CLTC:A.Sinha-NIPS'24} and achieves an $\mathcal{O}(T^{V_g})$ bound instead. Furthermore, COLDQ enhances the current best $\mathcal{O}(T^{\frac{1}{2}+V_x})$ dynamic regret \cite{CLTC:H.Guo-NIPS'22} to $\mathcal{O}(T^{\frac{1+V_x}{2}})$.
	
	\item For time-varying constraints and strongly convex loss functions, COLDQ improves the current best $\mathcal{O}(T^{\frac{1}{2}+V_x})$ dynamic regret and $\mathcal{O}(T^\frac{1}{2}(\log{T})^\frac{1}{2})$ hard constraint violation \cite{CLTC:H.Guo-NIPS'22} to $\mathcal{O}(T^{\frac{1+V_x}{2}})$ and $\mathcal{O}(T^{V_g})$.
	
	\item For fixed constraints and both convex and strongly-convex loss functions, COLDQ improves the current best $\mathcal{O}(\log{T})$ hard constraint violation  \cite{CLTC:H.Guo-NIPS'22} to $\mathcal{O}(1)$, while maintaining the $\mathcal{O}(T^\frac{1+V_x}{2})$ dynamic regret.
	
\end{itemize}

\section{Constrained Online Convex Optimization}
\label{sec:3}

We can consider the constrained OCO problem as an iterative game between a learner and the system over $T$ time slots. At each time $t$, the learner first selects a decision $\mathbf{x}_t$ from a known feasible set $\mathcal{X}\subseteq\mathbb{R}^p$. The loss function $f_t(\mathbf{x}):\mathbb{R}^p\to\mathbb{R}$ and the constraint function $\mathbf{g}_t(\mathbf{x})=[g_t^1(\mathbf{x}),\dots,g_t^N(\mathbf{x})]^\top:\mathbb{R}^p\to\mathbb{R}^N$ are then revealed to the learner, incurring a loss of $f_t(\mathbf{x}_t)$ and a constraint violation of $\mathbf{g}_t(\mathbf{x}_t)$. Both the loss function $f_t(\mathbf{x})$ and the constraint function $\mathbf{g}_t(\mathbf{x})$ are unknown a priori and are allowed to change arbitrarily over time.

The goal of the learner is to select from the feasible set an online decision sequence that minimizes the total accumulated loss under time-varying constraints. This gives rise to the following time-varying constrained OCO problem
\begin{align}
	\textbf{P}:\quad \min_{\{\mathbf{x}_t\in\mathcal{X}\}}\quad&\sum_{t=1}^Tf_t(\mathbf{x}_t)\notag\\
	\text{s.t.}\quad~~~ &\mathbf{g}_t(\mathbf{x}_t)\preceq\mathbf{0},\quad\forall{t}.\label{eq:ltc}
\end{align}
When $\mathbf{g}_t(\mathbf{x})=\mathbf{g}(\mathbf{x}),\forall{t}$, $\textbf{P}$ becomes the OCO problem with fixed constraints.

\subsection{Assumptions}
\label{sec:3.1}

We make some mild and common assumptions on $\mathcal{X}$, $f_t(\mathbf{x})$, and
$\mathbf{g}_t(\mathbf{x})$ in the constrained OCO literature.  

\begin{assumption}\label{asm:R}
	\textit{The feasible set $\mathcal{X}$ is convex and bounded, \ie $\exists{R}>0$,
		such that $\Vert\mathbf{x}-\mathbf{y}\Vert\le{R},\forall\mathbf{x},\mathbf{y}\in\mathcal{X}.$}
\end{assumption}

\begin{assumption}\label{asm:D}
	\textit{The loss functions are convex with bounded subgradient over $\mathcal{X}$, \ie $\exists{D}>0$, such that $\Vert\nabla{f}_t(\mathbf{x})\Vert\le{D},\forall\mathbf{x}\in\mathcal{X},\forall{t}$.}
\end{assumption}

\begin{assumption}\label{asm:G}
	\textit{The constraint functions are convex and bounded over $\mathcal{X}$, \ie $\exists{G}>0$, such that $|g_t^n(\mathbf{x})|\le{G},\forall\mathbf{x}\in\mathcal{X},\forall{t},\forall{n}$.}
\end{assumption}

Note that we do not require the commonly assumed Slater condition (or any of its relaxed version), on each of the constraint function at each time, \ie $\exists\tilde{\mathbf{x}}_t\in\mathcal{X}$ and $\delta>0$, such that $g_t^n(\tilde{\mathbf{x}}_t)<-\delta,\forall{t},\forall{n}$, \cite{SLTC:H.Yu-NIPS'17,DLTC:T.Chen-TSP'17,SLTC:H.Yu-JMLR'20,SLTC:X.Wei-SIGMETRICS'20,DTCOCO-TON}. The Slater condition, \ie the existence of a shared interior point assumption, excludes equality constraints that are common in many practical applications.

\subsection{Performance Metrics}

Finding an optimal solution to \textbf{P} is known to be impossible since the current information about $f_t(\mathbf{x})$ and $\mathbf{g}_t(\mathbf{x})$ is not available when selecting $\mathbf{x}_t$ at each time $t$. Instead, the OCO literature measures the performance of a constrained online algorithm, by comparing it with some solution benchmarks. There are two commonly used benchmarks. One is the \textit{fixed offline} solution benchmark $\mathbf{x}^\star\in\arg\min_{\mathbf{x}\in\mathcal{X}}\{\sum_{t=1}^Tf_t(\mathbf{x})|\mathbf{g}_t(\mathbf{x})\preceq\mathbf{0},\forall{t}\}$. The resulting \textit{static} regret is defined as
\begin{align}
	\text{REG}_\text{s}(T)\triangleq\sum_{t=1}^T\big[f_t(\mathbf{x}_t)-f_t(\mathbf{x}^\star)\big].\label{eq:sreg}
\end{align}
Another one is the \textit{dynamic online} solution benchmark $\mathbf{x}_t^\star\in\arg\min_{\mathbf{x}\in\mathcal{X}}\big\{f_t(\mathbf{x})|\mathbf{g}_t(\mathbf{x})\preceq\mathbf{0}\big\}$. The resulting \textit{dynamic} regret is defined as
\begin{align}
	\text{REG}_\text{d}(T)\triangleq\sum_{t=1}^T\big[f_t(\mathbf{x}_t)-f_t(\mathbf{x}_t^\star)\big].\label{eq:dreg}
\end{align}
The difference between the dynamic regret in (\ref{eq:dreg}) and the static regret in (\ref{eq:sreg}) can scale linearly with $T$, \ie $\text{REG}_\text{d}(T)-\text{REG}_\text{s}(T)=\mathcal{O}(T)$ \cite{OT:B.Omar-OR'15}. For a thorough analysis, in this work, we provide upper bounds on both the dynamic regret and the static regret. 

There are also two commonly used performance metrics to quantify how much the time-varying constraints (\ref{eq:ltc}) are violated. One is the \textit{soft} constraint violation defined as
\begin{align}
	\text{VIO}_\text{s}(T)\triangleq\sum_{n=1}^N\bigg[\sum_{t=1}^Tg_t^n(\mathbf{x}_t)\bigg]_+,\label{eq:svio}
\end{align}
where $[\cdot]_+$ is the projector onto the non-negative space. The  above soft constraint violation allows the violation at individual time slots to be \textit{compensated} over time. Another one is the \textit{hard} constraint violation defined as
\begin{align}
	\text{VIO}_\text{h}(T)\triangleq\sum_{n=1}^N\sum_{t=1}^T\big[g_t^n(\mathbf{x}_t)\big]_+.\label{eq:hvio}
\end{align}
This hard constraint violation does \textit{not} allow the violation at a time slot to be compensated by any other time slot. From the definitions of the soft and hard constraint violations in (\ref{eq:svio}) and (\ref{eq:hvio}), we readily have $\text{VIO}_\text{s}(T)\le\text{VIO}_\text{h}(T)$. In this work, we provide upper bounds on the hard constraint violation, which apply to the soft constraint violation as well.

\subsection{Variation Measures}

In the context of time-varying constrained OCO, it is desirable for an online algorithm to simultaneously achieve sublinear dynamic regret and sublinear constraint violation. This dual objective, however, can be intractable due to the adversarial variations of the losses and constraints. The performance guarantees of a constrained OCO algorithm are inherently linked to the temporal variations of both $\{f_t(\mathbf{x})\}_{t=1}^T$ and $\{\mathbf{g}_t(\mathbf{x})\}_{t=1}^T$. Therefore, it is necessary to quantify the dynamics of the underlying time-varying constrained OCO problem $\textbf{P}$.

There are two common variation measures in the literature. The first one measures the fluctuations in the dynamic online solution benchmark $\{\mathbf{x}_t^\star\}_{t=1}^T$, which is also referred to as the \textit{path length} \cite{DLTC:T.Chen-TSP'17,DLTC:X.Cao-TAC'19,CLTC:X.Yi-ICML'21,CLTC:H.Guo-NIPS'22,DLTC:Q.Liu-SIGMETRICS'22,DTCOCO-TON}, given by
\begin{align}
	\sum_{t=2}^T\Vert\mathbf{x}_t^\star-\mathbf{x}_{t-1}^\star\Vert=\mathcal{O}\big(T^{V_x}\big),\label{eq:Vx}
\end{align}
where $V_x\in[0,1]$ represents the time variability of the dynamic online solution benchmark.

The other one focuses on the fluctuations in the constraint functions $\{\mathbf{g}_t\}_{t=1}^T$ \cite{DLTC:T.Chen-TSP'17,DLTC:Q.Liu-SIGMETRICS'22,DTCOCO-TON}
\begin{align}
	\sum_{t=2}^T\max_{\mathbf{x}\in\mathcal{X}}\Vert\mathbf{g}_t(\mathbf{x})-\mathbf{g}_{t-1}(\mathbf{x})\Vert=\mathcal{O}\big(T^{V_g}\big),\label{eq:Vg}
\end{align}
where $V_g\in[0,1]$. Note that for the fixed offline solution benchmark, \ie $\mathbf{x}_t^\star=\mathbf{x}^\star,\forall{t}$, we have $V_x=0$. Similarly, for fixed constraint functions, \ie $\mathbf{g}_t(\mathbf{x})=\mathbf{g}(\mathbf{x}),\forall{t}$, we have $V_g=0$. 

\section{Constrained Online Learning with Doubly-bounded Queue (COLDQ)}

We present the COLDQ algorithm for solving $\textbf{P}$. In COLDQ, we introduce a novel doubly-bounded virtual queue and a new Lyapunov drift design, which will be shown to provide improved regret and constraint violation bounds.

\subsection{Doubly-Bounded Virtual Queue}

We introduce a novel virtual queue $Q_t^n$ to track the amount of violation for each time-varying constraint $n$. At the end of each time $t>1$, after observing the constraint function $\mathbf{g}_t(\mathbf{x})$, we update the virtual queue as:
\begin{align}
	Q_t^n=\max\big\{(1-\eta)Q_{t-1}^n+[g_t^n(\mathbf{x}_t)]_+,\gamma\big\},\label{eq:vq}
\end{align}
where $\eta\in(0,1)$ and $\gamma\in(0,\frac{G}{\eta})$ are two algorithm parameters. Our virtual queue updating rule (\ref{eq:vq}) includes an additional penalty term $-\eta Q_{t-1}^n$ to avoid the virtual queue from becoming excessively large. Furthermore, (\ref{eq:vq}) enforces a minimum virtual queue length $\gamma$ to prevent the constraint violation being overly large. In the following lemma, we show that without the Slater condition, (\ref{eq:vq}) leads to \textit{both} a lower bound and an upper bound on the virtual queue.\footnote{Existing constrained OCO works that adopt the virtual queue techniques can be divided into two groups: \citet{SLTC:H.Yu-NIPS'17,SLTC:X.Wei-SIGMETRICS'20,SLTC:H.Yu-JMLR'20,DLTC:Q.Liu-SIGMETRICS'22,DTCOCO-TON} bound the soft constraint violation by constructing a virtual queue that admits an upper bound only. \citet{CLTC:H.Guo-NIPS'22} bound the hard constraint violation by constructing a virtual queue that enforces a lower bound only. In contrast, our virtual queue construction yields both a lower and an upper bound. Together with a new Lyapunov drift analysis that leverages both bounds, COLDQ provides improved performance guarantees over the current best results.}

\begin{lemma}\label{lm:vq}
	\textit{Under Assumption~\ref{asm:G}, the virtual queue in (\ref{eq:vq}) has both a lower and an upper bound for each time~$t$ and each constraint $n$, given by}
	\begin{align}
		\gamma\le{Q}_t^n\le\frac{G}{\eta}.\label{eq:bd-vq}
	\end{align}
\end{lemma}

As shown in Lemma~\ref{lm:vq}, the parameter $\eta$ can be seen as a \textit{virtual} Slater constant for the constraints (\ref{eq:ltc}) in $\textbf{P}$. This means that the virtual queue upper bound is independent of the actual Slater constant. Furthermore, the parameter $\gamma$ ensures that the virtual queue length is always strictly positive. Our virtual queue updating rule (\ref{eq:vq}) leads to straightforward lower and upper bounds on the virtual queue itself. These virtual queue bounds, however, cannot be directly translated into a bound on the constraint violation. In the following section, we will establish a connection between our virtual queue and the hard constraint violation via a new Lyapunov-drift-based approach.

\subsection{Lyapunov Drift}

We define a new Lyapunov drift for each $t>1$ as
\begin{align}
	\Delta_{t-1}\triangleq\frac{1}{2}\sum_{n=1}^N(Q_t^n-\gamma)^2-\frac{1}{2}\sum_{n=1}^N(Q_{t-1}^n-\gamma)^2.\label{eq:drift}
\end{align}
Compared with the standard Lyapunov drift that uses the quadratic virtual queue as the Lyapunov function, each virtual queue $Q_t^n$ is penalized by its lower bound $\gamma$ in (\ref{eq:drift}). The subsequent lemma establishes an upper bound for $\Delta_{t-1}$, leveraging both the lower and upper bounds of $Q_t^n$ in (\ref{eq:bd-vq}).

\begin{lemma}\label{lm:drift}
	\textit{Under Assumption~\ref{asm:G}, the Lyapunov drift in (\ref{eq:drift}) is upper bounded for any $t>1$ by}
	\begin{align}
		\Delta_{t-1}&\le\sum_{n=1}^N{Q}_{t-1}^n[g_{t-1}^n(\mathbf{x}_t)]_+-\gamma\sum_{n=1}^N[g_t^n(\mathbf{x}_t)]_+\notag\\
		&\quad+\frac{G\sqrt{N}}{\eta}\max_{\mathbf{x}\in\mathcal{X}}\Vert\mathbf{g}_t(\mathbf{x})-\mathbf{g}_{t-1}(\mathbf{x})\Vert+2NG^2.\!\!\label{eq:bd-drift}
	\end{align}
\end{lemma}

The above Lyapunov drift upper bound comprises two key terms. The second term on the right-hand side (RHS) of (\ref{eq:bd-drift})  accounts for the hard constraint violation $\sum_{n=1}^N[g_t^n(\mathbf{x}_t)]_+$, scaled by the virtual queue lower bound $\gamma$. The third term on the RHS of (\ref{eq:bd-drift}) captures the fluctuation in the two adjacent constraint functions $\max_{\mathbf{x}\in\mathcal{X}}\Vert\mathbf{g}_t(\mathbf{x})-\mathbf{g}_{t-1}(\mathbf{x})\Vert$, scaled by the virtual queue upper bound $\frac{G}{\eta}$. These two terms are crucial for relating the hard constraint violation $\text{VIO}_\text{h}(T)$ to the constraint variation measure in (\ref{eq:Vg}), leading to improved performance bounds over the current-best results.

\subsection{Algorithm Intuition}

We solve the following per-slot optimization problem $\textbf{P}_t$ to determine the decision $\mathbf{x}_t$ at each time $t>1$
\begin{align*}
	\textbf{P}_t:~\min_{\mathbf{x}\in\mathcal{X}}~&\langle\nabla{f}_{t-1}(\mathbf{x}_{t-1}),\mathbf{x}-\mathbf{x}_{t-1}\rangle+\alpha_{t-1}\Vert\mathbf{x}-\mathbf{x}_{t-1}\Vert^2\notag\\
	&\quad+\sum_{n=1}^NQ_{t-1}^n[g_{t-1}^n(\mathbf{x})]_+
\end{align*}
where $\alpha_{t-1}>0$ is another algorithm parameter and is non-decreasing, \ie $\alpha_t\ge\alpha_{t-1},\forall{t}>1$. From the Lyapunov drift upper bound established in Lemma~\ref{lm:drift}, we can see the intuition behind solving $\textbf{P}_t$. Specifically, the objective is to greedily minimize the upper bound on the following \textit{drift plus penalty} term:
\begin{align*}
	\Delta_{t-1}+\langle\nabla{f}_{t-1}(\mathbf{x}_{t-1}),\mathbf{x}-\mathbf{x}_{t-1}\rangle+\alpha_{t-1}\Vert\mathbf{x}-\mathbf{x}_{t-1}\Vert^2.
\end{align*}
Note that the last two terms on the RHS of (\ref{eq:bd-drift}) are independent of $\mathbf{x}_t$, and second term is omitted in $\textbf{P}_t$ since $\mathbf{g}_t(\mathbf{x})$ is not available when choosing $\mathbf{x}_t$.

Minimizing the above penalty term $\langle\nabla{f}_{t-1}(\mathbf{x}_{t-1}),\mathbf{x}-\mathbf{x}_{t-1}\rangle+\alpha_{t-1}\Vert\mathbf{x}-\mathbf{x}_{t-1}\Vert^2$ itself is equivalent to performing the standard gradient descent $\mathbf{x}_{t-1}-\frac{1}{2\alpha_{t-1}}\nabla{f}_{t-1}(\mathbf{x}_{t-1})$. The optimal solution to $\textbf{P}_t$ depends on the amount of constraint violation induced by such gradient descent. If $g_{t-1}^n(\mathbf{x}_{t-1}-\frac{1}{2\alpha_{t-1}}\nabla{f}_{t-1}(\mathbf{x}_{t-1}))\le0,\forall{n}$, \ie the gradient descent does not incur any constraint violation, then $\mathbf{x}_t\in\arg\min_{\mathbf{x}\in\mathcal{X}}\{\mathbf{x}_{t-1}-\frac{1}{2\alpha_{t-1}}\nabla{f}_{t-1}(\mathbf{x}_{t-1})\}$ is the optimal solution to $\textbf{P}_t$. Otherwise, the gradient descent direction is shifted towards minimizing $Q_{t-1}^n[g_{t-1}^n(\mathbf{x}_t)]_+$ to reduce the constraint violation. The virtual queue $Q_{t-1}^n$ balances between loss minimization and violation reduction.

\subsection{The COLDQ Algorithm}
\label{sec:4.4}

In Algorithm~\ref{alg:1}, we summarize the proposed COLDQ algorithm. COLDQ consists of two main steps. The first step updates the decision variable $\mathbf{x}_t$ at the beginning of each time $t$ based on the gradient of the previous loss function $\nabla{f}_{t-1}(\mathbf{x}_{t-1})$ and the previous constraint function $\mathbf{g}_{t-1}(\mathbf{x})$. This primal update is designed to balance the accumulated loss minimization and the constraint violation control. The second step updates the virtual queue $Q_t^n,\forall{n}$ at the end of each $t$, after observing the constraint function $\mathbf{g}_t(\mathbf{x})$. This dual update is to track the amount of hard constraint violation. Note that COLDQ solves at each time $t$ a \textit{convex} optimization problem $\textbf{P}_t$, which can be efficiently solved in polynomial time. We will discuss the algorithm parameters $\alpha_t$, $\eta$, $\gamma$ to derive the best performance bounds for COLDQ in Section~\ref{sec:5.5}.\footnote{Using time-varying $\alpha_t$ allows COLDQ to match the current best static regret bound for strongly convex loss functions (see Corollary~\ref{cor:strong} in Section~\ref{sec:5.5}), and to incorporate the expert-tracking techniques to bound the dynamic regret without the knowledge of $V_x$ later (see the remark in Section~\ref{sec:5.5}).}

\begin{algorithm}[!t]
	\caption{\underline{C}onstrained \underline{O}nline \underline{L}earning with \underline{D}oubly-bounded \underline{Q}ueue (COLDQ)}
	\label{alg:1}
	\begin{algorithmic}[1]
		\STATE{Initialize non-decreasing sequence $\{\alpha_t\}\in(0,+\infty)$, $\eta\in(0,1)$,
			and $\gamma\in(0,\frac{G}{\eta})$. Choose $\mathbf{x}_1\in\mathcal{X}$
			arbitrarily and let $Q_1^n=\gamma,\forall n$.\\ At each time $t=2,\dots,T$,
			do the following:}
		
		\STATE{Update decision $\mathbf{x}_t$ by solving $\textbf{P}_t$.}
		\STATE{Observe $\nabla{f}_t(\mathbf{x}_t)$ and $\mathbf{g}_t(\mathbf{x})$.}
		\STATE{Update virtual queue $Q_t^n,\forall{n}$ via (\ref{eq:vq}).}
	\end{algorithmic}
	
\end{algorithm}

\section{Performance Bounds of COLDQ}
\label{sec:5}

\subsection{Preliminary Analysis}

The subsequent lemma establishes a per-slot performance guarantee of the COLDQ algorithm.

\begin{lemma}\label{lm:fgt}
	\textit{Under Assumptions \ref{asm:R}-\ref{asm:G}, the online decision sequence generated by COLDQ satisfies the following inequality for any $t>1$:}
	\begin{align}
		&\big[f_{t-1}(\mathbf{x}_{t-1})-f_{t-1}(\mathbf{x}_{t-1}^\star)\big]+\sum_{n=1}^NQ_{t-1}^n[g_{t-1}^n(\mathbf{x}_t)]_+\notag\\
		&\quad\le2R\alpha_{t-1}\Vert\mathbf{x}_t^\star-\mathbf{x}_{t-1}^\star\Vert+R^2(\alpha_t-\alpha_{t-1})+\frac{D^2}{4\alpha_{t-1}}\notag\\
		&\qquad+\big(\alpha_{t-1}\Vert\mathbf{x}_{t-1}^\star-\mathbf{x}_{t-1}\Vert^2-\alpha_t\Vert\mathbf{x}_t^\star-\mathbf{x}_t\Vert^2\big).\!\!\label{eq:bd-fgt}
	\end{align}
\end{lemma}

Lemma~\ref{lm:fgt} is the key to bridge the per-slot optimization problem $\textbf{P}_t$ and the performance bounds of COLDQ. From Lemma~\ref{lm:fgt}, we can \textit{separately} bound the dynamic regret and the hard constraint violation by substituting different lower bounds on $Q_{t-1}^n[g_{t-1}^n(\mathbf{x}_t)]_+$ into (\ref{eq:bd-fgt}). Note that adopting the soft constraint violation measure necessitates jointly bounding the regret and constraint violation.

\subsection{Bounding Dynamic Regret}
\label{sec:5.2}

The virtual queue length is always positive due to its lower bound, and the hard constraint violation is non-negative by definition. Hence, their product $Q_{t-1}^n[g_{t-1}^n(\mathbf{x}_t)]_+$ in (\ref{eq:bd-fgt}) is guaranteed to be non-negative. Unlike the analysis for soft-constrained OCO algorithms, this unique property enables us to bound the dynamic regret of COLDQ in the following theorem, without needing to explicitly consider the hard constraint violation.

\begin{theorem}\label{thm:reg}
	\textit{Under Assumptions~\ref{asm:R}-\ref{asm:G}, the dynamic regret of the COLDQ algorithm is upper bounded by}
	\begin{align}
		\text{REG}_\text{d}(T)&\le2R\sum_{t=2}^T\alpha_{t-1}\Vert\mathbf{x}_t^\star-\mathbf{x}_{t-1}^\star\Vert+\frac{D^2}{4}\sum_{t=1}^{T}\frac{1}{\alpha_t}\notag\\
		&\quad+R^2\alpha_T+DR.\label{eq:bd-reg}
	\end{align}
\end{theorem}

From Theorem~\ref{thm:reg}, we readily have an upper bound on the static regret $\text{REG}_\text{s}(T)$ by substituting $\mathbf{x}_t^\star=\mathbf{x}^\star,\forall{t}$ into the dynamic regret bound (\ref{eq:bd-reg}).

\subsection{Bounding Hard Constraint Violation}

The following theorem establishes a bound on the hard constraint violation incurred by the COLDQ algorithm. This is achieved by converting the term $Q_{t-1}^n[g_{t-1}^n(\mathbf{x}_t)]_+$ in (\ref{eq:bd-fgt}) to $[g_t^n(\mathbf{x}_t)]_+$ through the Lyapunov drift upper bound (\ref{eq:bd-drift}).

\begin{theorem}\label{thm:vio}
	\textit{Under Assumptions~\ref{asm:R}-\ref{asm:G}, the hard constraint violation of COLDQ is upper bounded by}
	\begin{align}
		&\text{VIO}_\text{h}(T)\le\frac{G\sqrt{N}}{\eta\gamma}\sum_{t=2}^T\max_{\mathbf{x}\in\mathcal{X}}\Vert\mathbf{g}_t(\mathbf{x})-\mathbf{g}_{t-1}(\mathbf{x})\Vert\notag\\
		&\qquad+\frac{2R}{\gamma}\sum_{t=2}^{T}\alpha_{t-1}\Vert\mathbf{x}_t^\star-\mathbf{x}_{t-1}^\star\Vert+\frac{D^2}{4\gamma}\sum_{t=1}^{T}\frac{1}{\alpha_t}\notag\\
		&\qquad+(DR+2NG^2)\frac{T}{\gamma}+R^2\frac{\alpha_T}{\gamma}+NG.\label{eq:bd-vio}
	\end{align}
\end{theorem}

To establish a hard constraint violation bound for fixed constraints, we can simply substitute $\mathbf{g}_t(\mathbf{x})=\mathbf{g}(\mathbf{x}),\forall{t}$ into the bound for time-varying constraints (\ref{eq:bd-vio}).

\subsection{Strongly Convex Case}
\label{sec:5.4}

We further consider the case of strongly convex loss functions as in \cite{CLTC:X.Yi-ICML'21,CLTC:H.Guo-NIPS'22,DCLTC:X.Yi-TAC'23}.
\begin{assumption}\label{asm:strong}
	\textit{The loss functions are $\mu$-strongly convex in $\mathcal{X}$ for some $\mu>0$ \ie $f_t(\mathbf{y})\ge{f}_t(\mathbf{x})+\langle\nabla{f}_t(\mathbf{x}),\mathbf{y}-\mathbf{x}\rangle+\mu\Vert\mathbf{y}-\mathbf{x}\Vert^2,\forall\mathbf{x},\mathbf{y}\in\mathcal{X},\forall{t}$.}
\end{assumption}

The following theorem provides a static regret bound for COLDQ with Assumption~\ref{asm:strong}.
\begin{theorem}\label{thm:strong}
	\textit{Under Assumptions~\ref{asm:R}-\ref{asm:strong}, the static regret of the COLDQ algorithm is upper bounded by}
	\begin{align}
		\text{REG}_\text{s}(T)&\le\sum_{t=2}^{T-1}\big(\alpha_t-\alpha_{t-1}-\mu\big)\Vert\mathbf{x}^\star-\mathbf{x}_t\Vert^2\notag\\
		&\quad+\frac{D^2}{4}\sum_{t=1}^{T}\frac{1}{\alpha_t}+(\alpha_1-\mu)R^2+DR.\label{eq:bd-reg-strong}
	\end{align}
\end{theorem}

\subsection{Regret and Constraint Violation Bounds}
\label{sec:5.5}

From Theorems~\ref{thm:reg}-\ref{thm:strong}, we can derive the following corollaries on the regret and constraint violation bounds of COLDQ.

\begin{corollary}[Convex Loss]\label{cor:conv}
	\textit{Under Assumptions~\ref{asm:R}-\ref{asm:G}, for any $V_x\in[0,1]$ and $V_g\in[0,1]$,  let $\alpha_t=t^\frac{1-V_x}{2}$, $\eta=T^{-1}$ and $\gamma=\epsilon T$, where $\epsilon\in(0,G)$, COLDQ achieves:}
	\begin{align}
		\text{REG}_\text{d}(T)=\mathcal{O}\big(T^\frac{1+V_x}{2}\big),\quad\text{VIO}_\text{h}(T)=\mathcal{O}\big(T^{V_g}\big).
	\end{align}
\end{corollary}

\begin{corollary}[Strongly Convex Loss]\label{cor:strong}
	\textit{Under Assumptions~\ref{asm:R}-\ref{asm:strong}, for any $V_g\in[0,1]$, let $\alpha_t=\mu{t}$, $\eta=T^{-1}$, and $\gamma=\epsilon{T}$, where $\epsilon\in(0,G)$, COLDQ achieves:}
	\begin{align}
		\text{REG}_\text{s}(T)=\mathcal{O}\big(\log{T}\big),\quad\text{VIO}_\text{h}(T)=\mathcal{O}\big(T^{V_g}\big).
	\end{align}
\end{corollary}

From Corollary~\ref{cor:conv}, we readily have a static regret bound $\text{REG}_\text{s}(T)=\mathcal{O}\big(T^\frac{1}{2}\big)$ by setting $V_x=0$, and a hard constraint violation bound $\text{VIO}_\text{h}(T)=\mathcal{O}(1)$ for fixed constraints by setting $V_g=0$. From Corollary~\ref{cor:strong}, we also have $\text{VIO}_\text{h}(T)=\mathcal{O}(1)$ for fixed constraints.

\begin{remark}
	The same $\mathcal{O}(T^\frac{1+V_x}{2})$ dynamic regret and $\mathcal{O}(T^{V_g})$ hard constraint violation in Corollary \ref{cor:conv} can be achieved without the knowledge of $V_x$ to set the algorithm parameter $\alpha_t$. In the Appendix, we extend the basic COLDQ algorithm with expert tracking, which can achieve the same performance bounds as COLDQ without any prior knowledge of the system dynamics.
\end{remark}

\section{Experiments}
\label{sec:6}

We conduct experiments to evaluate the performance of COLDQ for both time-varying and fixed constraints. In the Appendix, we provide all the algorithm parameters used in our experiments, and detailed problem settings of the application to online job scheduling.

\subsection{Experiment on Time-Varying Constraints}

Similar to the problem considered in~\cite{CLTC:H.Guo-NIPS'22,DCLTC:X.Yi-TAC'23}, we set the loss function as $f_t(\mathbf{x}) = \frac{1}{2}||\mathbf{H}_t\mathbf{x}-\mathbf{y}_t||^{2}$, where $\mathbf{H}_t\in \mathbb{R}^{4 \times 10}$, $\mathbf{x}\in\mathbb{R}^{10}$, and $\mathbf{y}_t\in \mathbb{R}^4$. Each element of $\mathbf{H}_t$ is uniformly distributed, \ie $H_t^{i,j} \sim U(-1,1),\forall{i}, j$. Each element of $\mathbf{y}_t$ is generated as $y_t^i = \sum_{j=1}^{10} H_t^{i,j} + \epsilon_i$, where $\epsilon_i$ follows a standard normal distribution. We set the constraint function as $\mathbf{g}_t(\mathbf{x}) = \mathbf{A}_t\mathbf{x} - \mathbf{b}_t$, where $\mathbf{A}_t\in \mathbb{R}^{2 \times 10}$ and $\mathbf{b}_t \in \mathbb{R}^2$, and $\mathcal{X}= \{\mathbf{x} \mid \mathbf{0}\preceq\mathbf{x}\preceq\mathbf{5}\}$. We generate $A_t^{i,j}\sim U(0,1),\forall{i,j}$ and $b_t^i\sim U(0,1),\forall{i}$.

\begin{figure}[t]
	\centering
	\includegraphics[width=1\linewidth]{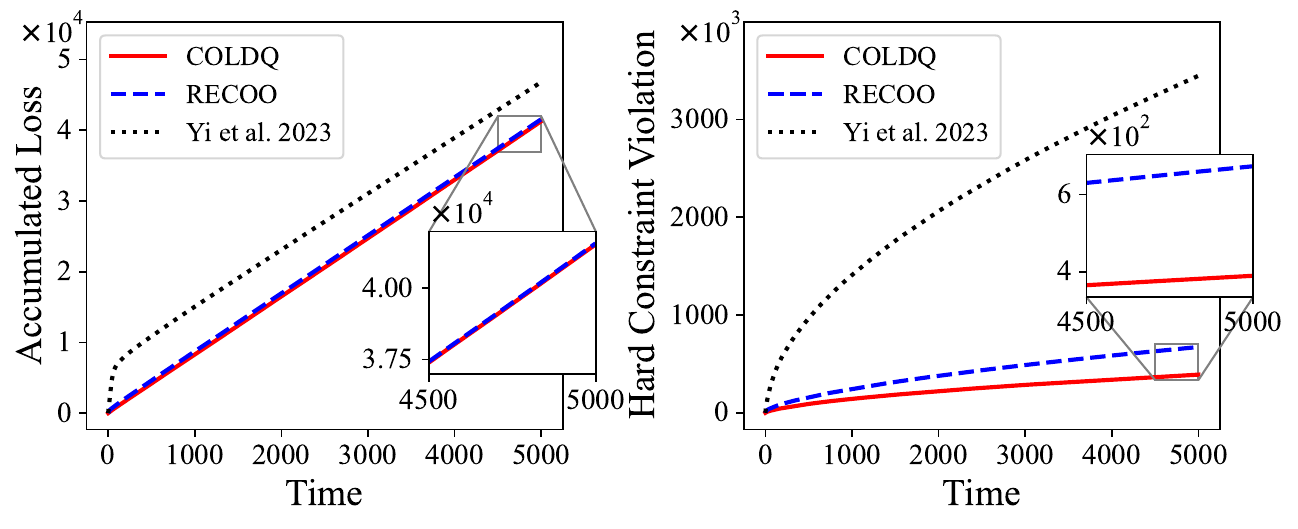}
	\caption{Experiment on time-varying constraints.}
	\label{fig:time-varying}
\end{figure}

We compare COLDQ with the state-of-the-art time-varying constrained OCO algorithms: RECOO~\cite{CLTC:H.Guo-NIPS'22} and Algorithm 1~\cite{DCLTC:X.Yi-TAC'23}. Fig~\ref{fig:time-varying} shows the accumulated loss and hard constraint violation. We can see that COLDQ achieves over $40\%$ lower constraint violation than RECOO without sacrificing the accumulated loss.

\begin{figure}[t]
	\centering
	\includegraphics[width=1\linewidth]{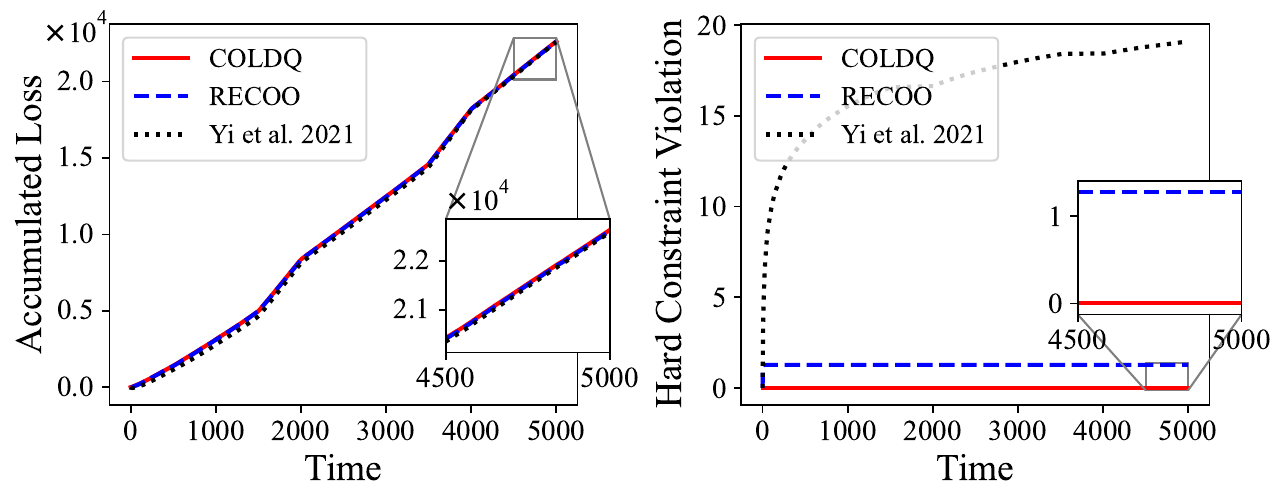}
	\caption{Experiment on online quadratic programming.}
	\label{fig:quadratic}
\end{figure}

\subsection{Experiment on Fixed Constraints}

We consider an online quadratic programming problem similar to \cite{CLTC:X.Yi-ICML'21}. We set the loss function as $f_t(\mathbf{x}) = ||\mathbf{x} - \boldsymbol{\theta}_t||^2 + 20\langle \boldsymbol{\theta}_t,\mathbf{x}\rangle$, where $\boldsymbol{\theta}_t=\boldsymbol{\theta}_t^1+\boldsymbol{\theta}_t^2+\boldsymbol{\theta}_t^3\in\mathbb{R}^{2}$ and $\mathbf{x}\in\mathbb{R}^2$. The time-varying parameters $\boldsymbol{\theta}_t$ are set as $\theta_t^{1,j}\sim U(-t^{1/10}, t^{1/10}),\forall{j}$; $\theta_t^{2,j}\sim U(-1, 0),\forall{j}$ for $t \in [1, 1500] \cup [2000, 3500] \cup [4000, 5000]$, and $\theta_t^{2,j} \sim U(0, 1),\forall{j}$ otherwise; and $\theta_t^{3,j} = (-1)^{\mu_t},\forall{j}$ with the sequence of $\mu_t$ being a random permutation of the vector $[1\colon5000]$.  We set the constraint function as $\mathbf{g}(\mathbf{x})=\mathbf{A}\mathbf{x}-\mathbf{b}$, where $\mathbf{A}\in\mathbb{R}^{3\times{2}}$ and $\mathbf{b}\in\mathbb{R}^{3}$ with $A_{i,j} \sim U(0.1, 0.5),\forall{i},j$, and $b_i \sim U(0, 0.3),\forall{i}$, and the feasible set as $\mathcal{X}= \{\mathbf{x} \mid \mathbf{0}\preceq\mathbf{x}\preceq\mathbf{1}\}$. We also experiment on the online linear programming problem considered in \cite{CLTC:X.Yi-ICML'21} and \cite{CLTC:H.Guo-NIPS'22}, by setting $f_t(\mathbf{x})=\langle \boldsymbol{\theta}_t, \mathbf{x} \rangle$ and keeping the rest of the problem settings unchanged.

\begin{figure}[t]
	\centering
	\includegraphics[width=1\linewidth]{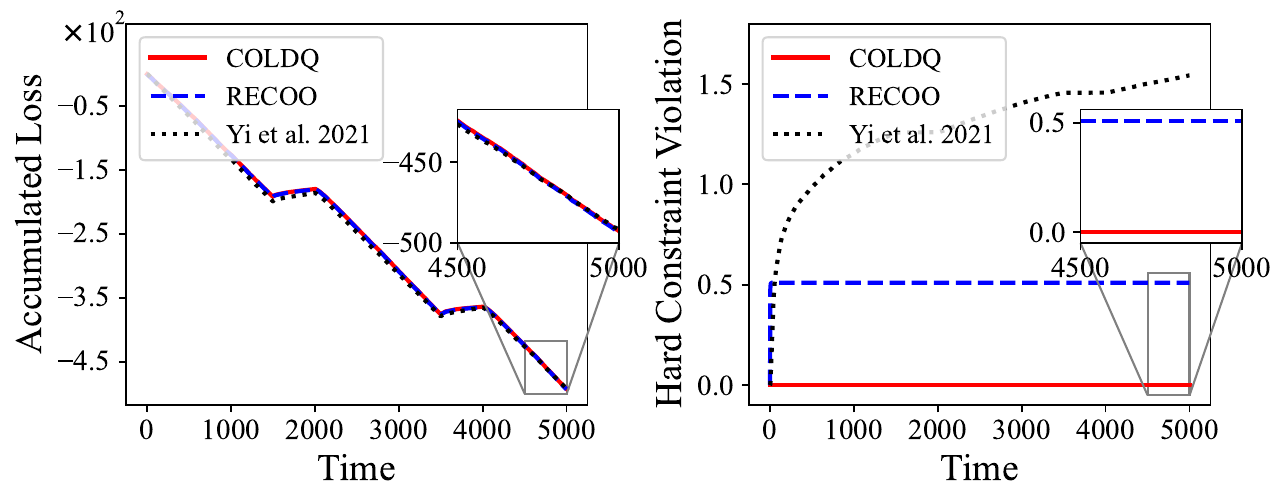}
	\caption{Experiment on online linear programming.}
	\label{fig:time-invariant}
\end{figure}

We compare COLDQ with the current-best time-invariant constrained OCO algorithms: Algorithm 1~\cite{CLTC:X.Yi-ICML'21} and RECOO \cite{CLTC:H.Guo-NIPS'22}. As shown in Figures~\ref{fig:quadratic} and~\ref{fig:time-invariant}, COLDQ demonstrates significant reductions in the hard constraint violation compared to RECOO~\cite{CLTC:H.Guo-NIPS'22}.

\subsection{Application to Online Job Scheduling}

\begin{figure}[t]
	\centering
	\includegraphics[width=1\linewidth]{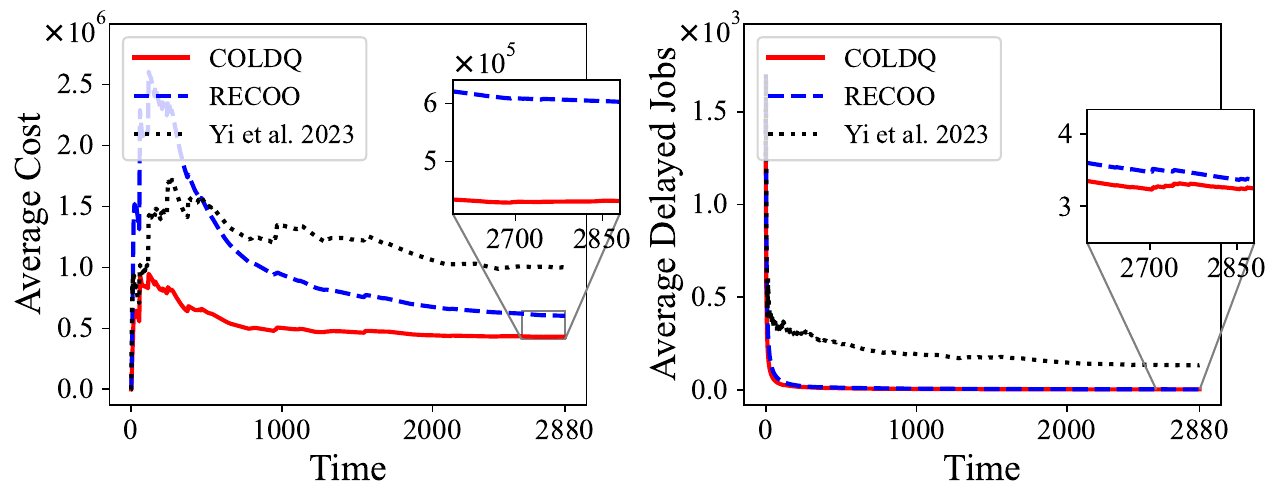}
	\caption{Experiment on online job scheduling.}
	\label{fig:job_schedule}
\end{figure}

We further apply COLDQ to online job scheduling using real-world datasets similar to \cite{SLTC:H.Yu-NIPS'17,CLTC:H.Guo-NIPS'22}. Figure~\ref{fig:job_schedule} shows the time-averaged energy cost and the number of delayed jobs. COLDQ demonstrates a significant reduction in the energy cost without compromising the service quality, compared with RECOO \cite{CLTC:H.Guo-NIPS'22} and Algorithm~1 \cite{DCLTC:X.Yi-TAC'23}.

\section{Conclusions}

We propose an effective COLDQ algorithm for OCO with time-varying constraints. We design a novel virtual queue that is bounded both from above and below to strictly control the hard constraint violation. Through a new Lyapunov drift analysis, COLDQ achieves $\mathcal{O}(T^\frac{1+V_x}{2})$ dynamic regret and $\mathcal{O}(T^{V_g})$ hard constraint violation. For the first time, the two bounds smoothly approach to the best-known $\mathcal{O}(T^\frac{1}{2})$ regret and $\mathcal{O}(1)$ violation, as the dynamics of the losses and constraints represented by $V_x$ and $V_g$ diminish. We further study the case of strongly-convex loss functions, and demonstrate that COLDQ matches the best-known $\mathcal{O}(\log{T})$ static regret while maintaining the $\mathcal{O}(T^{V_g})$ hard constraint violation. Moreover, we extend COLDQ with expert tracking capability, which allows it to achieve the same dynamic regret and hard constraint violation bounds without any prior knowledge of the system dynamics. Finally, experimental results complement our theoretical analysis.

\section*{Acknowledgments}
This work was supported in part by the Hong Kong Research Grants Council (RGC) Early Career Scheme (ECS) under grant 22200324.

\bibliography{aaai25}

\onecolumn
\section*{Appendix}
\appendix


\section{Auxiliary Lemmas}


\begin{lemma}[Lemma 2.8 \cite{BK-S.Shwartz'12}]\label{lm:strong}
	\textit{Let $\mathcal{X}$ be a convex set. Let $f(\mathbf{x}):\mathcal{X}\to\mathbb{R}$ be a $2\alpha$-strongly convex function with respect to a norm $\Vert\cdot\Vert$, and $\mathbf{x}^\circ\in\arg\min_{\mathbf{x}\in\mathcal{X}}f(\mathbf{x})$ be the optimal solution of $f(\mathbf{x})$. Then, we have $f(\mathbf{x}^\circ)\le{f}(\mathbf{x})-\alpha\Vert\mathbf{x}-\mathbf{x}^\circ\Vert^2,\forall\mathbf{x}\in\mathcal{X}$.}
\end{lemma}
\textit{Proof:} From the definition of strong convexity, we have  for any $\mathbf{x},\mathbf{y}\in\mathcal{X}$:
\begin{align*}
	f(\mathbf{x})\ge{f}(\mathbf{y})+\langle\nabla{f}(\mathbf{y}),\mathbf{x}-\mathbf{y}\rangle+\alpha\Vert\mathbf{x}-\mathbf{y}\Vert^2.
\end{align*}
Substituting $\mathbf{y}=\mathbf{x}^\circ$ into the above inequality and rearranging terms, we have for any $\mathbf{x}\in\mathcal{X}$:
\begin{align*}
	f(\mathbf{x}^\circ)\le{f}(\mathbf{x})-\langle\nabla{f}(\mathbf{x}^\circ),\mathbf{x}-\mathbf{x}^\circ\rangle-\alpha\Vert\mathbf{x}-\mathbf{x}^\circ\Vert^2.
\end{align*}
Further noting that the sufficient and necessary condition for $\mathbf{x}^\circ$ to be a global optimal solution of the convex function $f(\mathbf{x})$ is for any $\mathbf{x}\in\mathcal{X}$:
\begin{align*}
	\langle\nabla{f}(\mathbf{x}^\circ),\mathbf{x}-\mathbf{x}^\circ\rangle\ge0,
\end{align*}
we complete the proof.\hfill$\blacksquare$


\begin{lemma}[Time Series]\label{lm:series}
	\textit{The following time series is upper bounded for any $c\in[0,1)$ as}
	\begin{align*}
		\sum_{t=1}^T\frac{1}{t^c}\le\frac{1}{1-c}T^{1-c}.
	\end{align*}
\end{lemma}
\textit{Proof:} For any $c\in[0,1)$, we have
\begin{align*}
	\sum_{t=1}^T\frac{1}{t^c}\le\int_1^Tt^{-c}dt=\frac{1}{1-c}t^{1-c}\Big|_1^T=\frac{1}{1-c}(T^{1-c}-1)\le\frac{1}{1-c}T^{1-c}.
\end{align*}
\hfill$\blacksquare$


\begin{lemma}[Lemma 1 \cite{DReg:L.Zhang-NeurIPS'18}, Lemma 3 \cite{CLTC:X.Yi-ICML'21}]\label{lm:fm}
	\textit{Let $\mathcal{X}$ be a convex set. Let $\{l_t(\mathbf{x}):\mathcal{X}\to\mathbb{R}\}_{t=1}^T$ be a sequence of convex functions. Assume $l_t(\mathbf{x})$ is bounded, \ie $\exists{F}>0$, such that $|l_t(\mathbf{x})|\le{F},\forall\mathbf{x}\in\mathcal{X},\forall{t}$. Let $M\in\mathbb{N}_+$ and $\kappa>0$ be constants. Let $\{\mathbf{x}_t[m]\in\mathcal{X}\}_{t=1}^T,\forall{m}\in\mathcal{M}=\{1,\dots,M\}$ be $M$ sequences of decisions. Then, for any given $w_1[m]\in(0,1)$ that satisfies $\sum_{m=1}^Mw_1[m]=1$, let $\{\mathbf{x}_t\}_{t=1}^T$ be a sequence of decisions updated via
		\begin{align*}
			\mathbf{x}_t=\sum_{m=1}^Mw_t[m]\mathbf{x}_t[m]
		\end{align*}
		where
		\begin{align*}
			w_t[m]=\frac{w_{t-1}[m]e^{-\kappa{l}_{t-1}(\mathbf{x}_{t-1}[m])}}{\sum_{m=1}^Mw_{t-1}[m]e^{-\kappa{l}_{t-1}(\mathbf{x}_{t-1}[m])}}.
		\end{align*}
		Then, for any $T\ge1$, we have
		\begin{align*}
			\sum_{t=1}^Tl_t(\mathbf{x}_t)-\min_{m\in\mathcal{M}}\bigg\{\sum_{t=1}^Tl_t(\mathbf{x}_t[m])+\frac{1}{\kappa}\ln\bigg(\frac{1}{w_1[m]}\bigg)\bigg\}\le\frac{\kappa{F}^2T}{2}.
		\end{align*}
	}
\end{lemma}

\textit{Proof:} See the proof of Lemma~1 in \cite{DReg:L.Zhang-NeurIPS'18}.


\section{Proof of Lemma~\ref{lm:vq}}

\textit{Proof:} We first prove $Q_t^n\le\frac{G}{\eta},\forall{t},\forall{n}$ by induction. From the initialization of the virtual queue, we have $Q_1^n=\gamma\le\frac{G}{\eta},\forall{n}$. Then, suppose $Q_{\tau-1}^n\le\frac{G}{\eta},\forall{n}$ for some $\tau>1$, we now prove $Q_\tau^n\le\frac{G}{\eta},\forall{n}$. From the virtual queue updating rule 
(\ref{eq:vq}), we consider the following two cases:
\begin{itemize}
	\item If $(1-\eta)Q_{\tau-1}^n+[g_\tau^n(\mathbf{x}_\tau)]_+>\gamma$, we have
	\begin{align}
		Q_\tau^n&=(1-\eta)Q_{\tau-1}^n+[g_\tau^n(\mathbf{x}_\tau)]_+\le(1-\eta)Q_{\tau-1}^n+|g_\tau^n(\mathbf{x}_\tau)|\stackrel{(a)}{\le}(1-\eta)\frac{G}{\eta}+G=\frac{G}{\eta}
	\end{align}
	where $(a)$ follows from $Q_{\tau-1}^n\le\frac{G}{\eta}$ by induction
	and  the bound on $|g_t^n(\mathbf{x})|$ in Assumption~\ref{asm:G}.
	
	\item If $(1-\eta)Q_{\tau-1}^n+[g_\tau^n(\mathbf{x}_\tau)]_+\le\gamma$,
	we have
	\begin{align}
		Q_\tau^n=\gamma<\frac{G}{\eta}.
	\end{align}
\end{itemize}

Combining the above two cases, we have proven by induction that $Q_t^n\le\frac{G}{\eta},\forall{t},\forall{n}$.

Further noting that $Q_1^n=\gamma,\forall{n}$ from initialization and ${Q}_t^n\ge\gamma,\forall{t}>1,\forall{n}$ from (\ref{eq:vq}), we complete the proof.\hfill$\blacksquare$


\section{Proof of Lemma~\ref{lm:drift}}

\textit{Proof:} From the virtual queue updating rule in (\ref{eq:vq}), for any $t>1$ and $n$, we have
\begin{align}
	\frac{1}{2}(Q_t^n-\gamma)^2&=\frac{1}{2}\Big[\max\big\{(1-\eta)Q_{t-1}^n+[g_t^n(\mathbf{x}_t)]_+,\gamma\big\}-\gamma\Big]^2\stackrel{(a)}{\le}\frac{1}{2}\Big[(Q_{t-1}^n-\gamma)+\big([g_t^n(\mathbf{x}_t)]_+-\eta{Q}_{t-1}^n\big)\Big]^2\notag\\
	&=\frac{1}{2}(Q_{t-1}^n-\gamma)^2+\frac{1}{2}\big([g_t^n(\mathbf{x}_t)]_+-\eta{Q}_{t-1}^n\big)^2+Q_{t-1}^n[g_t^n(\mathbf{x}_t)]_+-\gamma[g_t^n(\mathbf{x}_t)]_+-\eta{Q}_{t-1}^n(Q_{t-1}^n-\gamma)\label{eq:bd-drift-1}
\end{align}
where $(a)$ follows from$|\max\{a,b\}-b|\le|a-b|,\forall{a},b\ge0$.

We now bound the RHS of (\ref{eq:bd-drift-1}). For the second term on the RHS of (\ref{eq:bd-drift-1}), we have\begin{align}
	\big|[g_t^n(\mathbf{x}_t)]_+-\eta{Q}_{t-1}^n\big|&\stackrel{(a)}{\le}[g_t^n(\mathbf{x}_t)]_++\eta{Q}_{t-1}^n\le|g_t^n(\mathbf{x}_t)|+\eta{Q}_{t-1}^n\stackrel{(b)}{\le}{G}+\eta\frac{G}{\eta}=2G\label{eq:bd-drift-2}
\end{align}
where $(a)$ follows from the triangle inequality, and $(b)$ is because of the bound on $|g_t^n(\mathbf{x})|$ in Assumption~\ref{asm:G} and the virtual queue upper bound in (\ref{eq:bd-vq}).

For the third term on the RHS of (\ref{eq:bd-drift-1}), we have
\begin{align}
	Q_{t-1}^n[g_t^n(\mathbf{x}_t)]_+&=Q_{t-1}^n[g_{t-1}^n(\mathbf{x}_t)]_++Q_{t-1}^n\big([g_t^n(\mathbf{x}_t)]_+-[g_{t-1}^n(\mathbf{x}_t)]_+\big)\notag\\
	&\stackrel{(a)}{\le}Q_{t-1}^n[g_{t-1}^n(\mathbf{x}_t)]_++\frac{G}{\eta}\big|[g_t^n(\mathbf{x}_t)]_+-[g_{t-1}^n(\mathbf{x}_t)]_+\big|\stackrel{(b)}{\le}Q_{t-1}^n[g_{t-1}^n(\mathbf{x}_t)]_++\frac{G}{\eta}\big|g_t^n(\mathbf{x}_t)-g_{t-1}^n(\mathbf{x}_t)\big|\label{eq:bd-drift-3}
\end{align}
where $(a)$ follows from the virtual queue upper bound in (\ref{eq:bd-vq}),
and $(b)$ is because $|[a]_+-[b]_+|\le|a-b|$.

For the last term on the RHS of (\ref{eq:bd-drift-1}), we have
\begin{align}
	-\eta{Q}_{t-1}^n(Q_{t-1}^n-\gamma)\le0\label{eq:bd-drift-4}
\end{align}
which follows form the virtual queue lower bound in (\ref{eq:bd-vq}).

Substituting (\ref{eq:bd-drift-2})-(\ref{eq:bd-drift-4}) into the RHS of
(\ref{eq:bd-drift-1}), for
any $t>1$ and $n$, we have
\begin{align}
	\frac{1}{2}(Q_t^n-\gamma)^2&\le\frac{1}{2}(Q_{t-1}^n-\gamma)^2+2G^2+Q_{t-1}^n[g_{t-1}^n(\mathbf{x}_t)]_++\frac{G}{\eta}\big|g_t^n(\mathbf{x}_t)-g_{t-1}^n(\mathbf{x}_t)\big|-\gamma[g_t^n(\mathbf{x}_t)]_+\label{eq:bd-drift-5}
\end{align}

Rearranging terms of (\ref{eq:bd-drift-5}), and summing over $n=1,\dots,N$, we have
\begin{align}
	\Delta_{t-1}&=\frac{1}{2}\sum_{n=1}^N(Q_t^n-\gamma)^2-\frac{1}{2}\sum_{n=1}^N(Q_{t-1}^n-\gamma)^2\notag\\
	&\le\sum_{n=1}^NQ_{t-1}^n[g_{t-1}^n(\mathbf{x}_t)]_+-\gamma\sum_{n=1}^N[g_t^n(\mathbf{x}_t)]_++\frac{G}{\eta}\sum_{n=1}^N\big|g_t^n(\mathbf{x}_t)-g_{t-1}^n(\mathbf{x}_t)\big|+2NG^{2}.\label{eq:bd-drift-6}
\end{align}

From (\ref{eq:bd-drift-6}) and  noting that $\sum_{n=1}^N\big|g_t^n(\mathbf{x}_t)-g_{t-1}^n(\mathbf{x}_t)\big|\le\sqrt{N}\max_{\mathbf{x}\in\mathcal{X}}\Vert\mathbf{g}_t(\mathbf{x})-\mathbf{g}_{t-1}(\mathbf{x})\Vert$, we complete the proof. \hfill$\blacksquare$


\section{Proof of Lemma~\ref{lm:fgt}}

\textit{Proof:} Note that $\langle\nabla{f}_{t-1}(\mathbf{x}_{t-1}),\mathbf{x}-\mathbf{x}_{t-1}\rangle$
is affine in $\mathbf{x}$ over $\mathcal{X}$, $\sum_{n=1}^NQ_{t-1}^n[g_{t-1}^n(\mathbf{x})]_+$ is convex in $\mathbf{x}$
over $\mathcal{X}$ since the maximum and linear combination of convex functions
are also convex, and $\alpha_{t-1}\Vert\mathbf{x}-\mathbf{x}_{t-1}\Vert^2$ is $2\alpha_{t-1}$-strongly convex in $\mathbf{x}$ over $\mathcal{X}$. Therefore, the objective function of $\textbf{P}_t$ is $2\alpha_{t-1}$-strongly convex in $\mathbf{x}$ over $\mathcal{X}$. Further noting that $\mathbf{x}_t$ is the optimal solution to $\textbf{P}_t$, applying Lemma~\ref{lm:strong} to $\textbf{P}_t$, we have
\begin{align}
	&\langle\nabla{f}_{t-1}(\mathbf{x}_{t-1}),\mathbf{x}_t-\mathbf{x}_{t-1}\rangle+\sum_{n=1}^NQ_{t-1}^n[g_{t-1}^n(\mathbf{x}_t)]_++\alpha_{t-1}\Vert\mathbf{x}_t-\mathbf{x}_{t-1}\Vert^2\notag\\
	&\quad\le\langle\nabla{f}_{t-1}(\mathbf{x}_{t-1}),\mathbf{x}_{t-1}^\star-\mathbf{x}_{t-1}\rangle+\sum_{n=1}^NQ_{t-1}^n[g_{t-1}^n(\mathbf{x}_{t-1}^\star)]_++\alpha_{t-1}\Vert\mathbf{x}_{t-1}^\star-\mathbf{x}_{t-1}\Vert^2-\alpha_{t-1}\Vert\mathbf{x}_{t-1}^\star-\mathbf{x}_t\Vert^2.\label{eq:fgt-1}
\end{align}

We now bound the RHS of (\ref{eq:fgt-1}). From the convexity of $f_{t-1}(\mathbf{x})$, the first term on the RHS of (\ref{eq:fgt-1}) is upper bounded by
\begin{align}
	\langle\nabla{f}_{t-1}(\mathbf{x}_{t-1}),\mathbf{x}_{t-1}^\star-\mathbf{x}_{t-1}\rangle\le{f}_{t-1}(\mathbf{x}_{t-1}^\star)-f_{t-1}(\mathbf{x}_{t-1}).\label{eq:fgt-2}
\end{align}

From the definition of the dynamic benchmark $\mathbf{x}_t^\star$, the second term on the RHS of (\ref{eq:fgt-1}) satisfies
\begin{align}
	\sum_{n=1}^NQ_{t-1}^n[g_{t-1}^n(\mathbf{x}_{t-1}^\star)]_+=0.\label{eq:fgt-3}
\end{align}

For the last two terms on the RHS of (\ref{eq:fgt-1}), we have
\begin{align}
	&\alpha_{t-1}\Vert\mathbf{x}_{t-1}^\star-\mathbf{x}_{t-1}\Vert^2-\alpha_{t-1}\Vert\mathbf{x}_{t-1}^\star-\mathbf{x}_t\Vert^2\notag\\
	&=\alpha_{t-1}\Vert\mathbf{x}_{t-1}^\star-\mathbf{x}_{t-1}\Vert^2-\alpha_t\Vert\mathbf{x}_t^\star-\mathbf{x}_t\Vert^2+\alpha_t\Vert\mathbf{x}_t^\star-\mathbf{x}_t\Vert^2-\alpha_{t-1}\Vert\mathbf{x}_{t-1}^\star-\mathbf{x}_t\Vert^2\notag\\
	&=\big(\alpha_{t-1}\Vert\mathbf{x}_{t-1}^\star-\mathbf{x}_{t-1}\Vert^2-\alpha_t\Vert\mathbf{x}_t^\star-\mathbf{x}_t\Vert^2\big)+\alpha_t\Vert\mathbf{x}_t^\star-\mathbf{x}_t\Vert^2-\alpha_{t-1}\Vert(\mathbf{x}_{t-1}^\star-\mathbf{x}_t^\star)+(\mathbf{x}_t^\star-\mathbf{x}_t)\Vert^2\notag\\
	&\stackrel{(a)}{\le}\big(\alpha_{t-1}\Vert\mathbf{x}_{t-1}^\star-\mathbf{x}_{t-1}\Vert^2-\alpha_t\Vert\mathbf{x}_t^\star-\mathbf{x}_t\Vert^2\big)+(\alpha_t-\alpha_{t-1})\Vert\mathbf{x}_t^\star-\mathbf{x}_t\Vert^2-\alpha_{t-1}\Vert\mathbf{x}_{t-1}^\star-\mathbf{x}_t^\star\Vert^2+2\alpha_{t-1}\Vert\mathbf{x}_t^\star-\mathbf{x}_t\Vert\Vert\mathbf{x}_t^\star-\mathbf{x}_{t-1}^\star\Vert\notag\\
	&\stackrel{(b)}{\le}\big(\alpha_{t-1}\Vert\mathbf{x}_{t-1}^\star-\mathbf{x}_{t-1}\Vert^2-\alpha_t\Vert\mathbf{x}_t^\star-\mathbf{x}_t\Vert^2\big)+(\alpha_t-\alpha_{t-1})R^2+2R\alpha_{t-1}\Vert\mathbf{x}_t^\star-\mathbf{x}_{t-1}^\star\Vert\label{eq:fgt-4}
\end{align}
where $(a)$ follows from $\Vert\mathbf{a}+\mathbf{b}\Vert^2\ge\Vert\mathbf{a}\Vert^2+\Vert\mathbf{b}\Vert^2-2\Vert\mathbf{a}\Vert\Vert\mathbf{b}\Vert$ and $(b)$ is because $\{\alpha_t\}$ being non-decreasing and $\mathcal{X}$ being bounded in Assumption~\ref{asm:R}.

Substituting (\ref{eq:fgt-2})-(\ref{eq:fgt-4}) into the RHS of (\ref{eq:fgt-1}) and rearranging terms, we have
\begin{align}
	&\big[f_{t-1}(\mathbf{x}_{t-1})-f_{t-1}(\mathbf{x}_{t-1}^\star)\big]+\sum_{n=1}^NQ_{t-1}^n[g_{t-1}^n(\mathbf{x}_t)]_+\notag\\
	&\quad\le2R\alpha_{t-1}\Vert\mathbf{x}_t^\star-\mathbf{x}_{t-1}^\star\Vert+(\alpha_t-\alpha_{t-1})R^2+\big(\alpha_{t-1}\Vert\mathbf{x}_{t-1}^\star-\mathbf{x}_{t-1}\Vert^2-\alpha_t\Vert\mathbf{x}_t^\star-\mathbf{x}_t\Vert^2\big)\notag\\
	&\qquad-\langle\nabla{f}_{t-1}(\mathbf{x}_{t-1}),\mathbf{x}_t-\mathbf{x}_{t-1}\rangle-\alpha_{t-1}\Vert\mathbf{x}_t-\mathbf{x}_{t-1}\Vert^2.\label{eq:fgt-5}
\end{align}

For the last two terms on the RHS of (\ref{eq:fgt-5}), completing the square, we have
\begin{align}
	&-\langle\nabla{f}_{t-1}(\mathbf{x}_{t-1}),\mathbf{x}_t-\mathbf{x}_{t-1}\rangle-\alpha_{t-1}\Vert\mathbf{x}_t-\mathbf{x}_{t-1}\Vert^2\notag\\
	&\quad=-\bigg\Vert\frac{\nabla{f}_{t-1}(\mathbf{x}_{t-1})}{2\sqrt{\alpha_{t-1}}}+\sqrt{\alpha_{t-1}}(\mathbf{x}_t-\mathbf{x}_{t-1})\bigg\Vert^2+\frac{\Vert\nabla{f}_{t-1}(\mathbf{x}_{t-1})\Vert^2}{4\alpha_{t-1}}\stackrel{(a)}{\le}\frac{D^2}{4\alpha_{t-1}}\label{eq:fgt-6}
\end{align}
where $(a)$ follows from the bound on $\nabla{f}_t(\mathbf{x})$ in Assumption~\ref{asm:D}.

Substituting (\ref{eq:fgt-6}) into (\ref{eq:fgt-5}) yields (\ref{eq:bd-fgt}).\hfill$\blacksquare$


\section{Proof of Theorem~\ref{thm:reg}}

\textit{Proof:} Summing (\ref{eq:bd-fgt}) over $t=2,\dots,T$ and noting that $Q_{t-1}^{n}[g_{t-1}^{n}(\mathbf{x}_t)]_+\ge0,\forall{t}>1,\forall{n}$, we have
\begin{align}
	&\sum_{t=1}^{T-1}\big[f_t(\mathbf{x}_t)-f_t(\mathbf{x}_t^\star)\big]\notag\\
	&\quad\le2R\sum_{t=2}^T\alpha_{t-1}\Vert\mathbf{x}_t^\star-\mathbf{x}_{t-1}^\star\Vert+R^2\sum_{t=2}^T(\alpha_t-\alpha_{t-1})+\frac{D^2}{4}\sum_{t=1}^{T-1}\frac{1}{\alpha_t}+\sum_{t=2}^T\big(\alpha_{t-1}\Vert\mathbf{x}_{t-1}^\star-\mathbf{x}_{t-1}\Vert^2-\alpha_t\Vert\mathbf{x}_t^\star-\mathbf{x}_t\Vert^2\big)\notag\\
	&\quad\le2R\sum_{t=2}^T\alpha_{t-1}\Vert\mathbf{x}_t^\star-\mathbf{x}_{t-1}^\star\Vert+R^2(\alpha_T-\alpha_1)+\frac{D^2}{4}\sum_{t=1}^T\frac{1}{\alpha_t}+\alpha_1\Vert\mathbf{x}_1^\star-\mathbf{x}_1\Vert^2-\alpha_T\Vert\mathbf{x}_T^\star-\mathbf{x}_T\Vert^2\notag\\
	&\quad\stackrel{(a)}{\le}2R\sum_{t=2}^T\alpha_{t-1}\Vert\mathbf{x}_t^\star-\mathbf{x}_{t-1}^\star\Vert+R^2(\alpha_T-\alpha_1)+\frac{D^2}{4}\sum_{t=1}^T\frac{1}{\alpha_t}+\alpha_1R^2\notag\\
	&\quad=2R\sum_{t=2}^T\alpha_{t-1}\Vert\mathbf{x}_t^\star-\mathbf{x}_{t-1}^\star\Vert+R^2\alpha_T+\frac{D^2}{4}\sum_{t=1}^T\frac{1}{\alpha_t}\label{eq:bd-reg-1}
\end{align}
where $(a)$ follows from $\mathcal{X}$ being bounded in Assumption~\ref{asm:R}.

Also, for any $t$, we have
\begin{align}
	f_t(\mathbf{x}_t)-f_t(\mathbf{x}_t^\star)&\stackrel{(a)}{\le}\langle\nabla{f}_t(\mathbf{x}_t^\star),\mathbf{x}_t^\star-\mathbf{x}_t\rangle\stackrel{(b)}{\le}\Vert\nabla{f}_t(\mathbf{x}_t^\star)\Vert\Vert\mathbf{x}_t^\star-\mathbf{x}_t\Vert\stackrel{(c)}{\le}DR\label{eq:bd-reg-2}
\end{align}
where $(a)$ follows from the convexity of $f_t(\mathbf{x})$, $(b)$ is due to the fact that $\langle\mathbf{a},\mathbf{b}\rangle\le\Vert\mathbf{a}\Vert\Vert\mathbf{b}\Vert$, and $(c)$ follows from $\mathcal{X}$ and $\nabla{f}_t(\mathbf{x})$ being bounded in Assumptions~\ref{asm:R} and \ref{asm:D} , respectively.

Combining (\ref{eq:bd-reg-1}) and $f_T(\mathbf{x}_T)-f_T(\mathbf{x}_T^\star)\le{DR}$ from (\ref{eq:bd-reg-2}), we complete the proof. \hfill$\blacksquare$


\section{Proof of Theorem~\ref{thm:vio}}

\textit{Proof:} Substituting the upper bound on the Lyapunov drift in (\ref{eq:bd-drift})
of Lemma~\ref{lm:drift} into the per-slot performance bound of COLDQ in (\ref{eq:bd-fgt}) of Lemma~\ref{lm:fgt} and
rearranging terms, we have
\begin{align}
	\gamma\sum_{n=1}^N[g_t^n(\mathbf{x}_t)]_+&\le\frac{G\sqrt{N}}{\eta}\max_{\mathbf{x}\in\mathcal{X}}\Vert\mathbf{g}_t(\mathbf{x})-\mathbf{g}_{t-1}(\mathbf{x})\Vert+2R\alpha_{t-1}\Vert\mathbf{x}_t^\star-\mathbf{x}_{t-1}^\star\Vert+\frac{D^2}{4\alpha_{t-1}}\notag\\
	&\quad+\big(\alpha_{t-1}\Vert\mathbf{x}_{t-1}^\star-\mathbf{x}_{t-1}\Vert^2-\alpha_t\Vert\mathbf{x}_t^\star-\mathbf{x}_t\Vert^2\big)+R^2(\alpha_t-\alpha_{t-1})\notag\\
	&\quad+\big[f_{t-1}(\mathbf{x}_{t-1}^\star)-f_{t-1}(\mathbf{x}_{t-1})\big]-\Delta_{t-1}+2NG^2.\label{eq:bd-vio-1}
\end{align}

Dividing both sides of (\ref{eq:bd-vio-1}) by $\gamma$ and summing over
$t=2,\dots,T$, we have
\begin{align}
	\sum_{n=1}^N\sum_{t=2}^T[g_t^n(\mathbf{x}_t)]_+&\le\frac{G\sqrt{N}}{\eta\gamma}\sum_{t=2}^T\max_{\mathbf{x}\in\mathcal{X}}\Vert\mathbf{g}_t(\mathbf{x})-\mathbf{g}_{t-1}(\mathbf{x})\Vert+\frac{2R}{\gamma}\sum_{t=2}^T\alpha_{t-1}\Vert\mathbf{x}_t^\star-\mathbf{x}_{t-1}^\star\Vert+\frac{D^2}{4\gamma}\sum_{t=2}^T\frac{1}{\alpha_{t-1}}\notag\\
	&\quad+\frac{1}{\gamma}\sum_{t=2}^T\big(\alpha_{t-1}\Vert\mathbf{x}_{t-1}^\star-\mathbf{x}_{t-1}\Vert^2-\alpha_t\Vert\mathbf{x}_t^\star-\mathbf{x}_t\Vert^2\big)+\frac{R^2}{\gamma}\sum_{t=2}^T(\alpha_t-\alpha_{t-1})\notag\\
	&\quad+\frac{1}{\gamma}\sum_{t=2}^{T}\big[f_{t-1}(\mathbf{x}_{t-1}^\star)-f_{t-1}(\mathbf{x}_{t-1})\big]-\frac{1}{\gamma}\sum_{t=2}^{T}\Delta_{t-1}+2NG^2\frac{T}{\gamma}.\label{eq:bd-vio-2}
\end{align}

We now bound the RHS of (\ref{eq:bd-vio-2}). We have
\begin{align}
	\sum_{t=2}^T\big(\alpha_{t-1}\Vert\mathbf{x}_{t-1}^\star-\mathbf{x}_{t-1}\Vert^2-\alpha_t\Vert\mathbf{x}_t^\star-\mathbf{x}_t\Vert^2\big)=\alpha_1\Vert\mathbf{x}_1^\star-\mathbf{x}_1\Vert^2-\alpha_T\Vert\mathbf{x}_T^\star-\mathbf{x}_T\Vert^2\le{R}^2\alpha_1\label{eq:bd-vio-3}
\end{align}
which follows from the bound on $\mathcal{X}$ in Assumption~\ref{asm:R}.

Also, we have
\begin{align}
	\sum_{t=2}^{T}\big[f_{t-1}(\mathbf{x}_{t-1}^\star)-f_{t-1}(\mathbf{x}_{t-1})\big]\le{DR}T\label{eq:bd-vio-4}
\end{align}
which follows from (\ref{eq:bd-reg-2}) in the proof of Theorem~\ref{thm:reg}.

We can show that
\begin{align}
	-\sum_{t=2}^{T}\Delta_{t-1}&\stackrel{(a)}{=}\sum_{t=2}^{T}\bigg[\frac{1}{2}\sum_{n=1}^N(Q_{t-1}^n-\gamma)^2-\frac{1}{2}\sum_{n=1}^N(Q_t^n-\gamma)^2\bigg]\notag\\
	&=\frac{1}{2}\sum_{n=1}^N(Q_{1}^n-\gamma)^2-\frac{1}{2}\sum_{n=1}^N(Q_T^n-\gamma)^2\le\frac{1}{2}\sum_{n=1}^N(Q_{1}^n-\gamma)^2\stackrel{(b)}{=}0\label{eq:bd-vio-5}
\end{align}
where $(a)$ follows from the definition of $\Delta_{t-1}$ in (\ref{eq:drift}), and $(b)$ is because $Q_1^n=\gamma,\forall{n}$ by initialization.

Substituting (\ref{eq:bd-vio-3})-(\ref{eq:bd-vio-5}) into (\ref{eq:bd-vio-2}), we have
\begin{align}
	\sum_{n=1}^N\sum_{t=2}^T[g_t^n(\mathbf{x}_t)]_+&\le\frac{G\sqrt{N}}{\eta\gamma}\sum_{t=2}^T\max_{\mathbf{x}\in\mathcal{X}}\Vert\mathbf{g}_t(\mathbf{x})-\mathbf{g}_{t-1}(\mathbf{x})\Vert+\frac{2R}{\gamma}\sum_{t=2}^T\alpha_{t-1}\Vert\mathbf{x}_t^\star-\mathbf{x}_{t-1}^\star\Vert+\frac{D^2}{4\gamma}\sum_{t=1}^T\frac{1}{\alpha_t}\notag\\
	&\quad+\frac{R^2}{\gamma}\alpha_1+\frac{R^2}{\gamma}(\alpha_T-\alpha_1)+(DR+2NG^2)\frac{T}{\gamma}.
\end{align}

Further noting that $[g_1^n(\mathbf{x}_1)]_+\le|g_1^n(\mathbf{x}_1)|\le{G},\forall{n}$ from Assumption~\ref{asm:G}, we complete the proof.\hfill$\blacksquare$


\section{Proof of Theorem~\ref{thm:strong}}

\textit{Proof:} Replacing the dynamic benchmark $\{\mathbf{x}_t^\star\}$ with the offline benchmark $\mathbf{x}^\star$ in (\ref{eq:fgt-1}) of the proof of Lemma~\ref{lm:fgt}, we have
\begin{align}
	&\langle\nabla{f}_{t-1}(\mathbf{x}_{t-1}),\mathbf{x}_t-\mathbf{x}_{t-1}\rangle+\sum_{n=1}^NQ_{t-1}^n[g_{t-1}^n(\mathbf{x}_t)]_++\alpha_{t-1}\Vert\mathbf{x}_t-\mathbf{x}_{t-1}\Vert^2\notag\\
	&\quad\le\langle\nabla{f}_{t-1}(\mathbf{x}_{t-1}),\mathbf{x}^\star-\mathbf{x}_{t-1}\rangle+\sum_{n=1}^NQ_{t-1}^n[g_{t-1}^n(\mathbf{x}^\star)]_++\alpha_{t-1}\Vert\mathbf{x}^\star-\mathbf{x}_{t-1}\Vert^2-\alpha_{t-1}\Vert\mathbf{x}^\star-\mathbf{x}_t\Vert^2.\label{eq:bd-strong-1}
\end{align}

We now bound the RHS of (\ref{eq:bd-strong-1}). From the $\mu$-strongly convexity of $f_{t-1}(\mathbf{x})$, the first term on the RHS of (\ref{eq:bd-strong-1}) can be upper bounded as
\begin{align}
	\langle\nabla{f}_{t-1}(\mathbf{x}_{t-1}),\mathbf{x}^\star-\mathbf{x}_{t-1}\rangle\le{f}_{t-1}(\mathbf{x}^\star)-f_{t-1}(\mathbf{x}_{t-1})-\mu\Vert\mathbf{x}^\star-\mathbf{x}_{t-1}\Vert^2.\label{eq:bd-strong-2}
\end{align}

From the definition of the offline benchmark $\mathbf{x}^\star$,
the second term on the RHS of (\ref{eq:bd-strong-1}) satisfies
\begin{align}
	\sum_{n=1}^NQ_{t-1}^n[g_{t-1}^n(\mathbf{x}^\star)]_+=0.\label{eq:bd-strong-3}
\end{align}

Substituting (\ref{eq:bd-strong-2}) and (\ref{eq:bd-strong-3}) into the RHS of (\ref{eq:bd-strong-1}) and rearranging terms, we have
\begin{align}
	&f_{t-1}(\mathbf{x}_{t-1})-f_{t-1}(\mathbf{x}^\star)\notag\\
	&~\le\big(\alpha_{t-1}-\mu\big)\Vert\mathbf{x}^\star-\mathbf{x}_{t-1}\Vert^2-\alpha_{t-1}\Vert\mathbf{x}^\star-\mathbf{x}_t\Vert^2-\sum_{n=1}^NQ_{t-1}^n[g_{t-1}^n(\mathbf{x}_t)]_+-\langle\nabla{f}_{t-1}(\mathbf{x}_{t-1}),\mathbf{x}_t-\mathbf{x}_{t-1}\rangle-\alpha_{t-1}\Vert\mathbf{x}_t-\mathbf{x}_{t-1}\Vert^2\notag\\
	&~\stackrel{(a)}{\le} \big(\alpha_{t-1}-\mu\big)\Vert\mathbf{x}^\star-\mathbf{x}_{t-1}\Vert^2-\alpha_{t-1}\Vert\mathbf{x}^\star-\mathbf{x}_t\Vert^2+\frac{D^2}{4\alpha_{t-1}}\label{eq:bd-strong-4}
\end{align}
where $(a)$ follows from $Q_{t-1}^n[g_{t-1}^n(\mathbf{x}_t)]_+\ge0,\forall{n}$ and (\ref{eq:fgt-6}) in the proof of Lemma~\ref{lm:fgt}.

Summing (\ref{eq:bd-strong-4}) over $t=2,\dots,T$, we have
\begin{align}
	\sum_{t=1}^{T-1}\big[f_t(\mathbf{x}_t)-f_t(\mathbf{x}^\star)\big]&\le\sum_{t=1}^{T-1}\big(\alpha_t-\mu\big)\Vert\mathbf{x}^\star-\mathbf{x}_t\Vert^2-\sum_{t=2}^T\alpha_{t-1}\Vert\mathbf{x}^\star-\mathbf{x}_t\Vert^2+\frac{D^2}{4}\sum_{t=1}^{T-1}\frac{1}{\alpha_t}\notag\\
	&=(\alpha_1-\mu)\Vert\mathbf{x}^\star-\mathbf{x}_1\Vert^2+\sum_{t=2}^{T-1}\big(\alpha_t-\alpha_{t-1}-\mu\big)\Vert\mathbf{x}^\star-\mathbf{x}_t\Vert^2-\alpha_{T-1}\Vert\mathbf{x}^\star-\mathbf{x}_T\Vert^2+\frac{D^2}{4}\sum_{t=1}^{T-1}\frac{1}{\alpha_t}\notag\\
	&\stackrel{(a)}{\le}(\alpha_1-\mu)R^2+\sum_{t=2}^{T-1}\big(\alpha_t-\alpha_{t-1}-\mu\big)\Vert\mathbf{x}^\star-\mathbf{x}_t\Vert^2+\frac{D^2}{4}\sum_{t=1}^{T}\frac{1}{\alpha_t}
\end{align}
where $(a)$ follows from $\mathcal{X}$ being bounded in Assumption~\ref{asm:R}.

Further noting that $f_T(\mathbf{x}_T)-f_T(\mathbf{x}^\star)\le{DR}$ similar to the proof of (\ref{eq:bd-reg-2}), we complete the proof.\hfill$\blacksquare$


\section{Proof of Corollary~\ref{cor:conv}}

\textit{Proof:} Substituting $\alpha_t=t^\frac{1-V_x}{2}$ into the dynamic regret bound (\ref{eq:bd-reg}) in Theorem~\ref{thm:reg}, we have
\begin{align}
	\text{REG}_\text{d}(T)&\le2R\sum_{t=2}^T(t-1)^\frac{1-V_x}{2}\Vert\mathbf{x}_t^\star-\mathbf{x}_{t-1}^\star\Vert+\frac{D^2}{4}\sum_{t=1}^{T}\frac{1}{t^\frac{1-V_x}{2}}+R^2T^\frac{1-V_x}{2}+DR\notag\\
	&\stackrel{(a)}{\le}2RT^\frac{1-V_x}{2}\sum_{t=2}^T\Vert\mathbf{x}_t^\star-\mathbf{x}_{t-1}^\star\Vert+\frac{D^{2}}{2(1+V_x)}T^{\frac{1+V_x}{2}}+R^2T^\frac{1-V_x}{2}+DR\notag\\
	&\stackrel{(b)}{=}\mathcal{O}\big(T^{\frac{1+V_x}{2}}\big)+\mathcal{O}\big(T^{\frac{1-V_x}{2}}\big)+\mathcal{O}(1)=\mathcal{O}\big(T^{\frac{1+V_x}{2}}\big)
\end{align}
where $(a)$ follows from Lemma~\ref{lm:series} and $(b)$ is because $\sum_{t=2}^T\Vert\mathbf{x}_t^\star-\mathbf{x}_{t-1}^\star\Vert=\mathcal{O}(T^{V_x})$ in (\ref{eq:Vx}).

Substituting $\alpha_t=t^\frac{1-V_x}{2}$, $\eta=\frac{1}{T}$ and $\gamma=\epsilon{T}$
into the hard constraint violation bound (\ref{eq:bd-vio}) in Theorem~\ref{thm:vio}, we
have
\begin{align}
	\text{VIO}_\text{h}(T)&\le\frac{G\sqrt{N}}{\epsilon}\sum_{t=2}^T\max_{\mathbf{x}\in\mathcal{X}}\Vert\mathbf{g}_t(\mathbf{x})-\mathbf{g}_{t-1}(\mathbf{x})\Vert+\frac{2RT^\frac{1-V_x}{2}}{\epsilon{T}}\sum_{t=2}^{T}\Vert\mathbf{x}_t^\star-\mathbf{x}_{t-1}^\star\Vert+\frac{D^2}{4\epsilon{T}}\sum_{t=1}^{T}\frac{1}{t^\frac{1-V_x}{2}}\notag\\
	&\quad+\frac{DR+2NG^2}{\epsilon}+\frac{R^2T^\frac{1-V_x}{2}}{\epsilon{T}}+NG\notag\\
	&\stackrel{(a)}{\le}\frac{G\sqrt{N}}{\epsilon}\sum_{t=2}^T\max_{\mathbf{x}\in\mathcal{X}}\Vert\mathbf{g}_t(\mathbf{x})-\mathbf{g}_{t-1}(\mathbf{x})\Vert+\frac{2R}{\epsilon}T^\frac{-1-V_x}{2}\sum_{t=2}^{T}\Vert\mathbf{x}_t^\star-\mathbf{x}_{t-1}^\star\Vert+\frac{D^{2}}{2\epsilon(1+V_x)}T^{\frac{V_x-1}{2}}\notag\\
	&\quad+\frac{DR+2NG^2}{\epsilon}+\frac{R^2}{\epsilon}T^\frac{-1-V_x}{2}+NG\notag\\
	&\stackrel{(b)}{=}\mathcal{O}\big(T^{V_g}\big)+\mathcal{O}\big(T^\frac{V_x-1}{2}\big)+\mathcal{O}\big(T^\frac{-1-V_x}{2}\big)+\mathcal{O}\big(1\big)=\mathcal{O}\big(T^{V_g}\big)
\end{align}
where $(a)$ follows from Lemma~\ref{lm:series} and $(b)$ is because $\sum_{t=2}^T\Vert\mathbf{x}_t^\star-\mathbf{x}_{t-1}^\star\Vert=\mathcal{O}(T^{V_x})$
in (\ref{eq:Vx}) and $\sum_{t=2}^T\max_{\mathbf{x}\in\mathcal{X}}\Vert\mathbf{g}_t(\mathbf{x})-\mathbf{g}_{t-1}(\mathbf{x})\Vert=\mathcal{O}(T^{V_g})$ in (\ref{eq:Vg}).\hfill$\blacksquare$


\section{Proof of Corollary~\ref{cor:strong}}

\textit{Proof:} Substituting $\alpha_t=\mu{t}$ into the static regret bound (\ref{eq:bd-reg-strong}) in Theorem~\ref{thm:strong}, we have
\begin{align}
	\text{REG}_\text{s}(T)&\le\sum_{t=2}^{T-1}\big[\mu{t}-\mu(t-1)-\mu\big]\Vert\mathbf{x}^\star-\mathbf{x}_t\Vert^2+\frac{D^2}{4\mu}\!\sum_{t=1}^{T}\frac{1}{t}+(\mu-\mu)R^2+DR\notag\\
	&=\frac{D^2}{4\mu}\!\sum_{t=1}^{T}\frac{1}{t}+DR\le\frac{D^2}{4\mu}\log{T}+DR=\mathcal{O}(\log{T})+\mathcal{O}(1)=\mathcal{O}(\log{T}).
\end{align}

Substituting $\alpha_t=\mu{t}$, $\eta=\frac{1}{T}$, and $\gamma=\epsilon{T}$ into (\ref{eq:bd-vio}) in Theorem~\ref{thm:vio} with the dynamic benchmark $\{\mathbf{x}_t^\star\}$ replaced by the offline benchmark $\mathbf{x}^\star$, we have
\begin{align}
	\text{VIO}_\text{h}(T)&\le\frac{G\sqrt{N}}{\epsilon}\sum_{t=2}^T\max_{\mathbf{x}\in\mathcal{X}}\Vert\mathbf{g}_t(\mathbf{x})-\mathbf{g}_{t-1}(\mathbf{x})\Vert+\frac{D^2}{4\mu\epsilon{T}}\sum_{t=1}^{T}\frac{1}{t}+\frac{DR+2NG^2}{\epsilon}+\frac{\mu{R}^2}{\epsilon}+NG\notag\\
	&\le\frac{G\sqrt{N}}{\epsilon}\sum_{t=2}^T\max_{\mathbf{x}\in\mathcal{X}}\Vert\mathbf{g}_t(\mathbf{x})-\mathbf{g}_{t-1}(\mathbf{x})\Vert+\frac{D^2\log{T}}{4\mu\epsilon{T}}+\frac{DR+2NG^2}{\epsilon}+\frac{\mu{R}^2}{\epsilon}+NG\notag\\
	&\stackrel{(a)}{=}\mathcal{O}\big(T^{V_g}\big)+\mathcal{O}\big(T^{-1}\log{T}\big)+\mathcal{O}\big(1)=\mathcal{O}\big(T^{V_g}\big)
\end{align}
where $(a)$ follows from $\sum_{t=2}^T\max_{\mathbf{x}\in\mathcal{X}}\Vert\mathbf{g}_t(\mathbf{x})-\mathbf{g}_{t-1}(\mathbf{x})\Vert=\mathcal{O}(T^{V_g})$
in (\ref{eq:Vg}).\hfill$\blacksquare$

\section{COLDQ with Expert Tracking}
\label{app:J}

We extend the basic COLDQ algorithm with expert tracking in Algorithm~\ref{alg:2}, which can achieve the same performance bounds as COLDQ without the knowledge of $V_x$ to set the algorithm parameter $\alpha_t$. The idea of expert tracking is to run multiple Algorithm~\ref{alg:1} in parallel, each tracks a different $V_x$, and then aggregate the decisions of different experts. 

The expert-tracking algorithm in \cite{DReg:L.Zhang-NeurIPS'18} is for OCO with \textit{short-term} constraints only, while the algorithm in \cite{CLTC:X.Yi-ICML'21} is for OCO with \textit{fixed} constraints. In contrast, Algorithm~\ref{alg:2} is the first \textit{time-varying} constrained OCO algorithm to provide $\mathcal{O}(T^\frac{1+V_x}{2})$ dynamic regret and $\mathcal{O}(T^{V_g})$ constraint violation that recover the best-known $\mathcal{O}(T^\frac{1}{2})$ regret and $\mathcal{O}(1)$ violation, without any prior knowledge of the system dynamics.

\begin{algorithm}[htb]
	\caption{COLDQ-Expert}
	\label{alg:2}
	\begin{algorithmic}[1]
		\STATE{Initialize $M\in\mathbb{N}_+$ and $\kappa\in(0,+\infty)$; non-decreasing sequences $\{\alpha_t[m]\}\in(0,+\infty),\forall{m}$,
			$\eta\in(0,1)$, $\gamma\in(0,\frac{G}{\eta})$, and $w_1[m]=\frac{M+1}{m(m+1)M},\forall{m}$. Choose $\mathbf{x}_1[m]\in\mathcal{X},\forall{m}$
			arbitrarily, and let $\mathbf{x}_1=\sum_{m=1}^Mw_1[m]\mathbf{x}_1[m]$ and $Q_1^n[m]=\gamma,\forall{n},\forall{m}$.\\ At each time $t=2,\dots,T$,
			do the following:}
		
		\STATE{Update expert decision $\mathbf{x}_t[m],\forall{m}$ by solving
			\begin{align*}
				&\textbf{P}_t[m]:~\min_{\mathbf{x}\in\mathcal{X}}~\langle\nabla{f}_{t-1}(\mathbf{x}_{t-1}[m]),\mathbf{x}-\mathbf{x}_{t-1}[m]\rangle+\alpha_{t-1}[m]\Vert\mathbf{x}-\mathbf{x}_{t-1}[m]\Vert^2+\sum_{n=1}^NQ_{t-1}^n[m][g_{t-1}^n(\mathbf{x})]_+.
			\end{align*}
		}
		\STATE{Update decision $\mathbf{x}_t=\sum_{m=1}^Mw_t[m]\mathbf{x}_t[m]$.}
		\STATE{Observe $\nabla{f}_t(\mathbf{x}_t)$ and $\mathbf{g}_t(\mathbf{x})$.}
		\STATE{Update virtual queue $Q_t^n[m],\forall{n},\forall{m}$ via
			\begin{align*}
				Q_t^n[m]=\max\big\{(1-\eta)Q_{t-1}^n[m]+[g_t^n(\mathbf{x}_t[m])]_+,\gamma\big\}.\!\!\!\label{eq:vqm}
			\end{align*}
		}
		\STATE{Update} weight $w_{t+1}[m]=\frac{w_t[m]e^{-\kappa{l}_t(\mathbf{x}_t[m])}}{\sum_{m=1}^Mw_t[m]e^{-\kappa{l}_t(\mathbf{x}_t[m])}}$,
		where $l_t(\mathbf{x})=\langle\nabla{f}_t(\mathbf{x}_t),\mathbf{x}-\mathbf{x}_t\rangle$.
	\end{algorithmic}
\end{algorithm}

\begin{corollary}[Expert Tracking]\label{cor:expert}
	\textit{Under Assumptions~\ref{asm:R}-\ref{asm:G}, for any $V_x\in[0,1]$ and $V_g\in[0,1]$, let $M=\lfloor\frac{1}{2}\log_2(1+T)\rfloor+1$. $\kappa=T^{-\frac{1}{2}}$, $\alpha_t[m]=t^\frac{1}{2}/2^{m-1}$, $\eta=T^{-\frac{3}{2}}$ and $\gamma=\epsilon{T}^\frac{3}{2}$, where $\epsilon\in(0,G)$, COLDQ-Expert achieves:}
	\begin{align}
		\text{REG}_\text{d}(T)=\mathcal{O}\big(T^\frac{1+V_x}{2}\big),\quad\text{VIO}_\text{h}(T)=\mathcal{O}\big(T^{V_g}\big).
	\end{align}
\end{corollary}


\section{Proof of Corollary~\ref{cor:expert}}

\textit{Proof:} We first derive the dynamic regret bound. Following the Proof of (\ref{eq:bd-reg-1}) in Theorem~\ref{thm:reg}, we can show that for each $m\in\mathcal{M}$:
\begin{align}
	\sum_{t=1}^T\big[f_t(\mathbf{x}_t[m])-f_t(\mathbf{x}_t^\star)\big]&\le2R\sum_{t=2}^T\alpha_{t-1}[m]\Vert\mathbf{x}_t^\star-\mathbf{x}_{t-1}^\star\Vert+\frac{D^2}{4}\sum_{t=1}^{T}\frac{1}{\alpha_t[m]}+R^2\alpha_T[m]+DR\notag\\
	&\stackrel{(a)}{=}\frac{2R}{2^{m-1}}\sum_{t=2}^Tt^\frac{1}{2}\Vert\mathbf{x}_t^\star-\mathbf{x}_{t-1}^\star\Vert+\frac{D^22^{m-1}}{4}\sum_{t=1}^{T}t^{-\frac{1}{2}}+\frac{R^2T^\frac{1}{2}}{2^{m-1}}+DR\notag\\
	&\stackrel{(b)}{\le}4R\frac{T^\frac{1}{2}\sum_{t=2}^T\Vert\mathbf{x}_t^\star-\mathbf{x}_{t-1}^\star\Vert}{2^m}+\frac{D^2}{2}T^\frac{1}{2}2^{m-1}+2R^2\frac{T^\frac{1}{2}}{2^m}+DR\label{eq:expert-1}
\end{align}
where $(a)$ is because of setting $\alpha_t[m]=t^\frac{1}{2}/2^{m-1}$and $(b)$ follows from Lemma~\ref{lm:series}.

Since the number of experts is set as $M=\lfloor\frac{1}{2}\log_2(1+T)\rfloor+1$, there exists an expert
\begin{align}
	\tilde{m}=\Big\lfloor\frac{1}{2}\log_2\Big(1+\frac{\sum_{t=2}^T\Vert\mathbf{x}_t^\star-\mathbf{x}_{t-1}^\star\Vert}{R}\Big)\Big\rfloor+1\le{M}
\end{align}
such that
\begin{align}
	2^{\tilde{m}-1}\le\bigg(1+\frac{\sum_{t=2}^T\Vert\mathbf{x}_t^\star-\mathbf{x}_{t-1}^\star\Vert}{R}\bigg)^\frac{1}{2}\le2^{\tilde{m}}\label{eq:expert-2}.
\end{align}

Substituting (\ref{eq:expert-2}) into (\ref{eq:expert-1}), we have
\begin{align}
	\sum_{t=1}^{T}\big[f_t(\mathbf{x}_t[\tilde{m}])-f_t(\mathbf{x}_t^\star)\big]&\le4R\bigg(\frac{RT\big[\sum_{t=2}^T\Vert\mathbf{x}_t^\star-\mathbf{x}_{t-1}^\star\Vert\big]^2}{R+\sum_{t=2}^T\Vert\mathbf{x}_t^\star-\mathbf{x}_{t-1}^\star\Vert}\bigg)^\frac{1}{2}+\frac{D^2}{2}\bigg(T+\frac{T\sum_{t=2}^T\Vert\mathbf{x}_t^\star-\mathbf{x}_{t-1}^\star\Vert}{R}\bigg)^\frac{1}{2}\notag\\
	&\quad+2R^2\bigg(\frac{RT}{R+\sum_{t=2}^T\Vert\mathbf{x}_t^\star-\mathbf{x}_{t-1}^\star\Vert}\bigg)^\frac{1}{2}+DR\label{eq:expert-3}.
\end{align}

Also, we have
\begin{align}
	\sum_{t=1}^{T}\big[f_t(\mathbf{x}_t)-f_t(\mathbf{x}_t[\tilde{m}])\big]&\stackrel{(a)}{\le}\sum_{t=1}^{T}\big\langle\nabla{f}_t(\mathbf{x}_t),\mathbf{x}_t-\mathbf{x}_t[\tilde{m}]\big\rangle\notag\\
	&\stackrel{(b)}{=}\sum_{t=1}^T\big[l_t(\mathbf{x}_t)-l_t(\mathbf{x}_t[\tilde{m}])\big]\notag\\
	&\stackrel{(c)}{\le}\frac{1}{\kappa}\ln\bigg(\frac{1}{w_1[\tilde{m}]}\bigg)+\frac{\kappa{D}^2R^2T}{2}\label{eq:expert-4}.
\end{align}
where $(a)$ follows from the convexity of $f_t(\mathbf{x})$, $(b)$ is because $l_t(\mathbf{x})=\langle\nabla{f}_t(\mathbf{x}_t),\mathbf{x}-\mathbf{x}_t\rangle$, $(c)$ follows from Lemma~\ref{lm:fm} and $|l_t(\mathbf{x})|\le{DR},\forall\mathbf{x}\in\mathcal{X},\forall{t}$ under Assumptions~\ref{asm:R} and \ref{asm:D}.

Recall $w_1[m]=\frac{M+1}{m(m+1)M}$ by initialization, we have
\begin{align}
	\ln\bigg(\frac{1}{w_1[\tilde{m}]}\bigg)\le\ln\big(\tilde{m}(\tilde{m}+1)\big)\le2\ln(\tilde{m}+1)=2\ln\bigg(\Big\lfloor\frac{1}{2}\log_2\Big(1+\frac{\sum_{t=2}^T\Vert\mathbf{x}_t^\star-\mathbf{x}_{t-1}^\star\Vert}{R}\Big)\Big\rfloor+2\bigg).\label{eq:expert-5}
\end{align}

Substituting (\ref{eq:expert-5}) into (\ref{eq:expert-4}) and noting that $\kappa=T^{-\frac{1}{2}}$, we have
\begin{align}
	\sum_{t=1}^{T}\big[f_t(\mathbf{x}_t)-f_t(\mathbf{x}_t[\tilde{m}])\big]\le2\ln\bigg(\Big\lfloor\frac{1}{2}\log_2\Big(1+\frac{\sum_{t=2}^T\Vert\mathbf{x}_t^\star-\mathbf{x}_{t-1}^\star\Vert}{R}\Big)\Big\rfloor+2\bigg)T^\frac{1}{2}+\frac{\kappa{D}^2R^2}{2}T^\frac{1}{2}.\label{eq:expert-6}
\end{align}

Combining (\ref{eq:expert-3}) and (\ref{eq:expert-6}), we have
\begin{align}
	\text{REG}_\text{d}(T)&=\sum_{t=1}^{T}\big[f_t(\mathbf{x}_t)-f_t(\mathbf{x}_t^\star)\big]=\sum_{t=1}^{T}\big[f_t(\mathbf{x}_t)-f_t(\mathbf{x}_t[\tilde{m}])\big]+\sum_{t=1}^{T}\big[f_t(\mathbf{x}_t[\tilde{m}])-f_t(\mathbf{x}_t^\star)\big]\notag\\
	&\le4R\bigg(\frac{RT\big[\sum_{t=2}^T\Vert\mathbf{x}_t^\star-\mathbf{x}_{t-1}^\star\Vert\big]^2}{R+\sum_{t=2}^T\Vert\mathbf{x}_t^\star-\mathbf{x}_{t-1}^\star\Vert}\bigg)^\frac{1}{2}+\frac{D^2}{2}\bigg(T+\frac{T\sum_{t=2}^T\Vert\mathbf{x}_t^\star-\mathbf{x}_{t-1}^\star\Vert}{R}\bigg)^\frac{1}{2}+2R^2\bigg(\frac{RT}{R+\sum_{t=2}^T\Vert\mathbf{x}_t^\star-\mathbf{x}_{t-1}^\star\Vert}\bigg)^\frac{1}{2}\notag\\
	&\quad+2\ln\bigg(\Big\lfloor\frac{1}{2}\log_2\Big(1+\frac{\sum_{t=2}^T\Vert\mathbf{x}_t^\star-\mathbf{x}_{t-1}^\star\Vert}{R}\Big)\Big\rfloor+2\bigg)T^\frac{1}{2}+\frac{\kappa{D}^2R^2}{2}T^\frac{1}{2}+DR\notag\\
	&\stackrel{(a)}{=}\mathcal{O}\big(T^\frac{1+V_x}{2}\big)+\mathcal{O}\big(T^\frac{1}{2}\big)+\mathcal{O}\big(1\big)=\mathcal{O}\big(T^\frac{1+V_x}{2}\big)\label{eq:expert-7}
\end{align}
where $(a)$ follows from $\sum_{t=2}^T\Vert\mathbf{x}_t^\star-\mathbf{x}_{t-1}^\star\Vert=\mathcal{O}(T^{V_x})$
in (\ref{eq:Vx}).

We now derive the hard constraint violation bound. Replacing $\{\mathbf{x}_t\}$ with $\{\mathbf{x}_t[m]\}$ in the proof of Theorem~\ref{thm:vio}, for each expert $m$, we can show that
\begin{align}
	\sum_{n=1}^N\sum_{t=1}^T\big[g_t^n(\mathbf{x}_t[m])\big]_+&\le\frac{G\sqrt{N}}{\eta\gamma}\sum_{t=2}^T\max_{\mathbf{x}\in\mathcal{X}}\Vert\mathbf{g}_t(\mathbf{x})-\mathbf{g}_{t-1}(\mathbf{x})\Vert+\frac{2R}{\gamma}\sum_{t=2}^{T}\alpha_{t-1}[m]\Vert\mathbf{x}_t^\star-\mathbf{x}_{t-1}^\star\Vert+\frac{D^2}{4\gamma}\sum_{t=1}^{T}\frac{1}{\alpha_t[m]}\notag\\
	&\quad+(DR+2NG^2)\frac{T}{\gamma}+R^2\frac{\alpha_T[m]}{\gamma}+NG\notag\\
	&\stackrel{(a)}{=}\frac{G\sqrt{N}}{\epsilon}\sum_{t=2}^T\max_{\mathbf{x}\in\mathcal{X}}\Vert\mathbf{g}_t(\mathbf{x})-\mathbf{g}_{t-1}(\mathbf{x})\Vert+\frac{2R}{\epsilon2^{m-1}}{T}^{-\frac{3}{2}}\sum_{t=2}^T(t-1)^\frac{1}{2}\Vert\mathbf{x}_t^\star-\mathbf{x}_{t-1}^\star\Vert\notag\\
	&\quad+\frac{D^22^{m-1}}{4\epsilon}T^{-\frac{3}{2}}\sum_{t=1}^Tt^{-\frac{1}{2}}+\frac{DR+2NG^2}{\epsilon}T^{-\frac{1}{2}}+\frac{R^2}{\epsilon2^{m-1}}T^{-1}+NG\notag\\
	&\stackrel{(b)}{\le}\frac{G\sqrt{N}}{\epsilon}\sum_{t=2}^T\max_{\mathbf{x}\in\mathcal{X}}\Vert\mathbf{g}_t(\mathbf{x})-\mathbf{g}_{t-1}(\mathbf{x})\Vert+\frac{2R}{\epsilon}{T}^{-1}\sum_{t=2}^T\Vert\mathbf{x}_t^\star-\mathbf{x}_{t-1}^\star\Vert+\frac{D^2}{2\epsilon}(1+T)^\frac{1}{2}T^{-1}\notag\\
	&\quad+\frac{DR+2NG^2}{\epsilon}T^{-\frac{1}{2}}+\frac{R^2}{\epsilon}T^{-1}+NG\notag\\
	&\stackrel{(c)}{=}\mathcal{O}\big(T^{V_g}\big)+\mathcal{O}\big(T^{V_x-1}\big)+O\big(1\big)=\mathcal{O}\big(T^{V_g}\big)\label{eq:expert-8}
\end{align}
where $(a)$ follows from setting $\alpha_t[m]=t^\frac{1}{2}/2^{m-1}$, $\eta=T^{-\frac{3}{2}}$, and $\gamma=\epsilon{T}^\frac{3}{2}$, $(b)$ is because $m\le\lfloor\frac{1}{2}\log_2(1+T)\rfloor+1,\forall{m}$ and $\sum_{t=1}^Tt^{-\frac{1}{2}}\le2T^\frac{1}{2}$ from Lemma~\ref{lm:series}, and $(c)$ follows from   $\sum_{t=2}^T\Vert\mathbf{x}_t^\star-\mathbf{x}_{t-1}^\star\Vert=\mathcal{O}(T^{V_x})$ in (\ref{eq:Vx}) and $\sum_{t=2}^T\max_{\mathbf{x}\in\mathcal{X}}\Vert\mathbf{g}_t(\mathbf{x})-\mathbf{g}_{t-1}(\mathbf{x})\Vert=\mathcal{O}(T^{V_g})$ in (\ref{eq:Vg}).

Noting that $g_t^n(\mathbf{x}),\forall{t},\forall{n}$ is convex and $\mathbf{x}_t=\sum_{m=1}^Mw_t[m]\mathbf{x}_t[m]$ with $\sum_{m=1}^Mw_t[m]=1,\forall{t}$, We then have
\begin{align}
	\sum_{n=1}^N\sum_{t=1}^T\big[g_t^n(\mathbf{x}_t)\big]_+&\stackrel{(a)}{=}\sum_{n=1}^N\sum_{t=1}^T\bigg[g_t^n\bigg(\sum_{m=1}^Mw_t[m]\mathbf{x}_t[m]\bigg)\bigg]_+\notag\\
	&\stackrel{(b)}{\le}\sum_{n=1}^N\sum_{t=1}^T\sum_{m=1}^Mw_t[m]\big[g_t^n(\mathbf{x}_t[m])\big]_+\notag\\
	&\le\sum_{n=1}^N\sum_{t=1}^T\sum_{m=1}^M\big[g_t^n(\mathbf{x}_t[m])\big]_+\stackrel{(c)}{=}\mathcal{O}\big(T^{V_g}\big)\label{eq:expert-9}
\end{align}
where $(a)$ follows from $\mathbf{x}_t=\sum_{m=1}^Mw_t[m]\mathbf{x}_t[m],\forall{t}$, $(b)$ is due to the convexity of $g_t^n(\mathbf{x}),\forall{t},\forall{n}$, and $\sum_{m=1}^Mw_t[m]=1,\forall{t}$, and $(c)$ follows from $\sum_{n=1}^N\sum_{t=1}^T\big[g_t^n(\mathbf{x}_t[m])\big]_+=\mathcal{O}(T^{V_g}),\forall{m}$ in (\ref{eq:expert-8}). \hfill$\blacksquare$

\section{Experiment Details}
\label{app:L}

We provide all the algorithm parameters used in our experiments, the intuition behind the fine-tuning of all algorithms for fair comparison, and the detailed problem settings of the application to online job scheduling.

\subsubsection{Experiment on Time-Varying Constraints.}

In Table~\ref{tab:time-varying-params}, we summarize all algorithm parameters used to generate Figure~\ref{fig:time-varying}. For fair comparison among COLDQ (this work), RECOO \cite{CLTC:H.Guo-NIPS'22}, and Algorithm 1 \cite{DCLTC:X.Yi-TAC'23}, we fine-tuned their suggested parameters such that they reach as close accumulated loss as possible at the end. In this way, we can focus on comparing the algorithm performance in terms of the hard constraint violation.

\begin{table}[htb]
	\renewcommand{\arraystretch}{1.2}
	\caption{Algorithm parameters for the experiment on time-varying constraints.}
	\label{tab:time-varying-params}
	\centering
	\begin{tabular}{c|c}
		\hline
		Algorithm & Parameters \\ \hline \hline
		COLDQ (this work) & $\alpha_{t}=t^{1/2}$, $\gamma=\epsilon{T}$ , $\eta=1/T$, $\epsilon=0.5$ \\ \hline
		RECOO \cite{CLTC:H.Guo-NIPS'22} & $\alpha_{t}=0.5t^{1/2}$, $\gamma_{t}=t^{1/2+\epsilon}$, $\eta_{t}=t^{1/2}$, $\epsilon=0.01$ \\ \hline
		Algorithm 1 \cite{DCLTC:X.Yi-TAC'23} & $\alpha_{t}=3/t^{1/2}$, $\gamma_{t}=0.1/t^{1/2}$, $\beta_{t}=5/t^{1/2}$ \\ \hline
	\end{tabular}
\end{table}

\subsubsection{Experiment on Fixed Constraints.}

In Tables~\ref{tab:quadratic-params} and \ref{tab:time-invariant-params}, we summarize all algorithm parameters used to generate Figures~\ref{fig:quadratic} and \ref{fig:time-invariant}. Similar to the previous experiment, we fine-tuned the suggested parameters of all algorithms to reach nearly the same accumulated loss, and compare their hard constraint violations.

\begin{table}[htb]
	\renewcommand{\arraystretch}{1.2}
	\caption{Algorithm parameters for the experiment on online quadratic programming.}
	\label{tab:quadratic-params}
	\centering
	\begin{tabular}{c|c}
		\hline
		Algorithm & Parameters \\ \hline \hline
		COLDQ (this work) & $\alpha_{t}=t^{1/2}$, $\gamma=\epsilon{T}$, $\eta=1/T$, $\epsilon=0.5$ \\ \hline
		RECOO \cite{CLTC:H.Guo-NIPS'22} & $\alpha_{t}=t^{1/2}$, $\gamma_{t}=7t^{1/2+\epsilon}$, $\eta_{t}=t^{1/2}$, $\epsilon=0.01$ \\ \hline
		Algorithm 1 \cite{CLTC:X.Yi-ICML'21} & $\alpha_{t}=100/T^{1/2}$, $\gamma_{t}=2.5/T^{c/2}$, $c=0.5$ \\ \hline
	\end{tabular}
\end{table}

\begin{table}[htb]
	\renewcommand{\arraystretch}{1.2}
	\caption{Algorithm parameters for the experiment on online linear programming.}
	\label{tab:time-invariant-params}
	\centering
	\begin{tabular}{c|c}
		\hline
		Algorithm & Parameters \\ \hline \hline
		COLDQ (this work) & $\alpha_{t}=t^{1/2}$, $\gamma=\epsilon{T}$, $\eta=1/T$, $\epsilon=0.5$ \\ \hline
		RECOO \cite{CLTC:H.Guo-NIPS'22} & $\alpha_{t}=t^{1/2}$, $\gamma_{t}=t^{1/2+\epsilon}$, $\eta_{t}=t^{1/2}$, $\epsilon=0.01$ \\ \hline
		Algorithm 1 \cite{CLTC:X.Yi-ICML'21} & $\alpha_{t}=2/T^{1/2}$, $\gamma_{t}=1/T^{c/2}$, $c=0.5$ \\ \hline
	\end{tabular}
\end{table}

\subsubsection{Application to Online Job Scheduling.}

We experiment on the practical online job schedulig problem considered in \cite{SLTC:H.Yu-NIPS'17,CLTC:H.Guo-NIPS'22}. This application aims at allocating power across data centers to minimize the energy cost subject to service quality constraints. We consider $100$ data centers equally distributed in $10$ regions. The duration of each time slot $t$ is 5 minutes. Let $\mathbf{x}_t=[x_t^1,\dots,x_t^{100}]^\top\in\mathbb{R}^{100}$ be the power allocation decision at time~$t$. The loss function $f_t(\mathbf{x}_t)$ representing the energy cost is set as $f_t(\mathbf{x}_t)=\langle\mathbf{c}_t,\mathbf{x}_t\rangle$, where $\mathbf{c}_t\in\mathbb{R}^{100}$ is the time-varying electricity price vector. The constraint function $g_t(\mathbf{x}_t)$ representing the service quality is set as $g_t(\mathbf{x}_t)=\lambda_t-\sum_{i=1}^{100}h_i(\mathbf{x}_t^i)$, where $\lambda_t$ is the job arrival rate and $h_i(x_t^i) = 4\log(1 + 4x_t^i)$ models the service capacity of each data center. The constraint violation measures the amount of delayed jobs not finished in time. We use real-world electricity price data from NYISO (available from http://www.nyiso.com/.) for 10 New York City regions between $05/01/2017$ and $05/10/2017$. The number of arriving jobs $\lambda_t$ at each time $t$ is generated from a Poisson distribution with mean $2500$. The feasible set $\mathcal{X}$ is set as $\mathcal{X}=\{\mathbf{x}_t|0\le{x}_t^i\le1000,\forall{t},\forall{i}\}$.

In Table~\ref{tab:job-scheduling-params}, we summarize all algorithm parameters used to generate Figure~\ref{fig:job_schedule}. We fine-tuned the suggested parameters of COLD (this work) and RECOO \cite{CLTC:H.Guo-NIPS'22} to reach nearly the same number of average delayed jobs at the end. We optimized the suggested parameters of Algorithm 1 \cite{DCLTC:X.Yi-TAC'23} to reach its best performance.

\begin{table}[!h]
	\renewcommand{\arraystretch}{1.2}
	\caption{Algorithm parameters for the application to online job scheduling.}
	\label{tab:job-scheduling-params}
	\centering
	\begin{tabular}{c|c}
		\hline
		Algorithm & Parameters \\ \hline \hline
		COLDQ (this work) & $\alpha_{t}=t^{1/2}$, $\gamma=\epsilon{T}$, $\eta=1/T$, $\epsilon=0.5$ \\ \hline
		RECOO \cite{CLTC:H.Guo-NIPS'22} & $\alpha_{t}=0.7t^{1/2}$, $\gamma_{t}=t^{1/2+\epsilon}$, $\eta_{t}=t^{1/2}$, $\epsilon=0.01$ \\ \hline
		Algorithm 1 \cite{DCLTC:X.Yi-TAC'23} & $\alpha_{t}=100/t^{1/2}$, $\gamma_{t}=1/t^{1/2}$, $\beta_{t}=1/t^{1/2}$ \\ \hline
	\end{tabular}
\end{table}

We implemented all algorithms in Python 3.8.19 with CVXPY 1.5.1. A laptop with Intel(R) Core(TM) i5-13600K CPU@3.50GHz and 32 GB of RAM can finish a single run of each algorithm within 10 ms per iteration.

\end{document}